\documentclass[11pt]{article}

\usepackage[preprint]{acl}

\usepackage{times}
\usepackage{latexsym}

\usepackage[T1]{fontenc}
\usepackage{hyphenat}

\usepackage[utf8]{inputenc}

\usepackage{microtype}
\usepackage{lipsum}

\usepackage{inconsolata}

\usepackage{graphicx}

\usepackage{booktabs}
\usepackage{multirow}
\usepackage[table]{xcolor}
\usepackage{hyperref}
\usepackage{enumitem}
\usepackage{todonotes}
\usepackage{pifont}

\newcommand{\red}[1]{\textcolor{red}{#1}}

\newcommand{\green}[1]{\textcolor{green}{#1}}
\newcommand{\redcell}[0]{\cellcolor{red!25}}
\newcommand{\blueell}[0]{\cellcolor{blue!20}}
\newcommand{\greencell}[0]{\cellcolor{green!25}}

\newcommand{\purplecell}[0]{\cellcolor{purple!20}}
\newcommand{\yellowcell}[0]{\cellcolor{yellow!20}}

\newcommand{\bold}[1]{\textbf{#1}}
\newcommand{\indoculturedialoguetitle}{\textbf{\textsc{CultureTalk-ID}}}

\title{\indoculturedialoguetitle: A Multi-Task Dialogue Benchmark for Cultural Commonsense in Indonesian Local Languages}

\author{Muhammad Dehan Al Kautsar\textsuperscript{1} \ Salsabila Pranida\textsuperscript{1} \\ \textbf{Bilal Elbouardi\textsuperscript{1}} \ \textbf{Fajri Koto\textsuperscript{1}} \\
\textsuperscript{1}Mohamed bin Zayed University of Artificial Intelligence \\
    \texttt{\small muhammad.dehan@mbzuai.ac.ae 
    } 
}

\begin{document}
\maketitle
\begin{abstract}
Culture is lived through conversation, yet existing Indonesian cultural commonsense benchmarks evaluate LLMs on short and isolated prompts, stripping away the dialogic context in which cultural nuances actually surface. We introduce \indoculturedialoguetitle, the first dialogue-based benchmark for cultural commonsense in Indonesian and its local languages, comprising 4,496 culturally grounded dialogues across 11 languages and 13 culturally salient topics, curated through a multi-stage human pipeline with native speakers to ensure authenticity. \indoculturedialoguetitle\ introduces three complementary tasks, namely dialogue-based multiple-choice cultural commonsense reasoning, culturally faithful machine translation, and language steering, which jointly probe whether LLMs can understand, transfer, and generate culturally grounded language.
Our results show that open-source models still underperform, particularly in the generation task and local language settings, highlighting a gap in their ability to capture and generate culturally grounded knowledge in conversational contexts.\footnote{The dataset link is anonymized.}

\end{abstract}

\begin{figure*}[t]
  \centering
  \includegraphics[width=1.9\columnwidth]{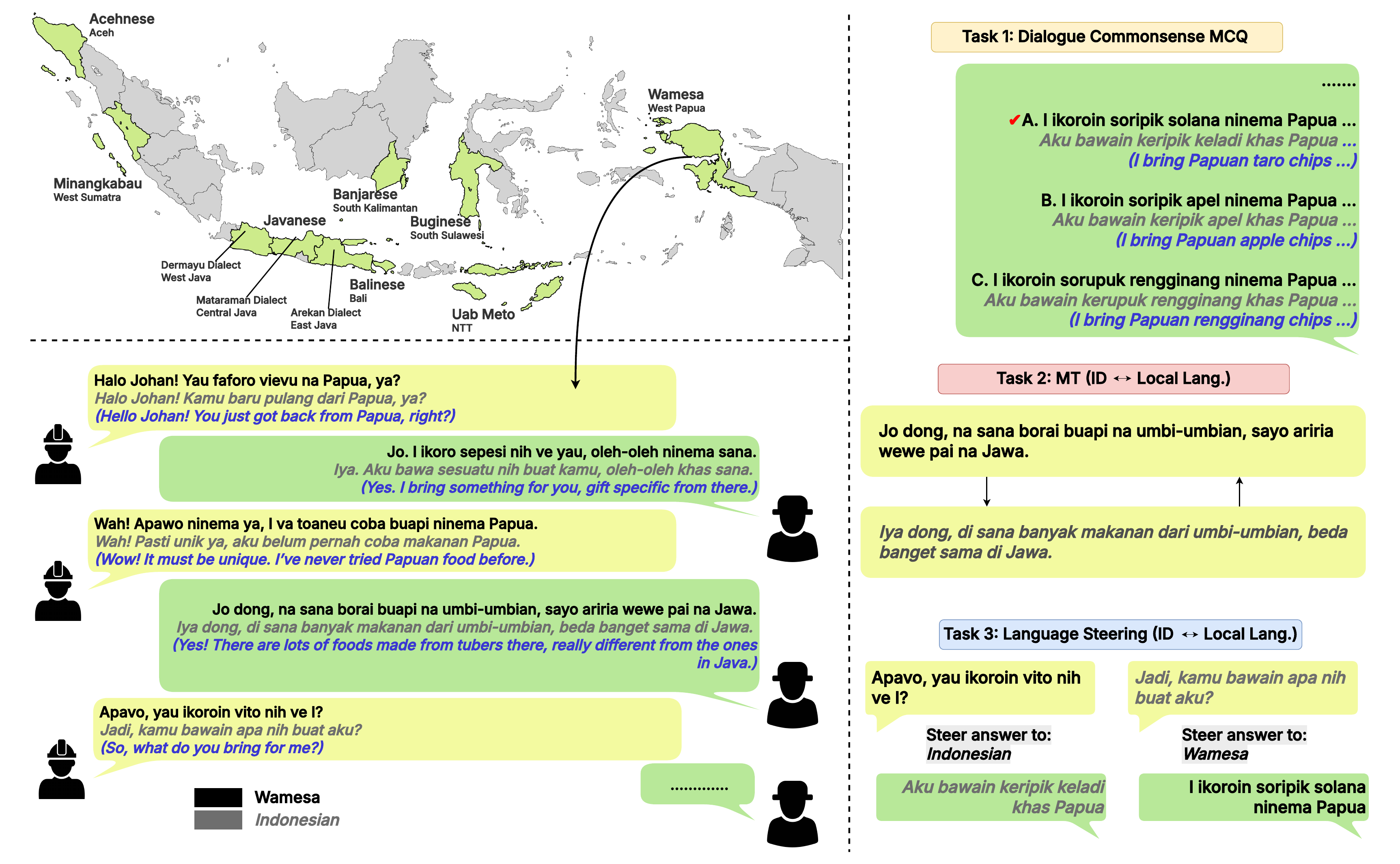}
  \caption{Overview of the \indoculturedialoguetitle\ benchmark. \indoculturedialoguetitle\ supports three tasks: (1) dialogue-based multiple-choice questions (MCQs) for cultural commonsense reasoning, (2) bidirectional machine translation between Indonesian and local languages, and (3) language steering.}
  \label{fig:indoculturedial-overview}
\end{figure*}

\section{Introduction}


Culture is expressed through conversation. It encompasses the patterns of thinking, communication, and behavior, along with material artifacts, that shape a community’s way of life and are transmitted across generations~\cite{liu-etal-2025-culturally, nguyen-2022-bayesian-mindsponge}. This transmission is inherently dynamic, as it is continually negotiated and renewed through everyday interactions between individuals and groups, emerging in conversations across languages and communities~\cite{suraj-2024-culture}. As LLMs are increasingly deployed as conversational agents for culturally diverse users, their ability to recognize and reason about culture within dialogue becomes critical, not only for producing contextually appropriate responses~\cite{wu-etal-2025-socialcc}, but also for capturing the implicit, community-specific norms and expectations that guide interaction, often referred to as cultural commonsense.

However, despite this inherently interactive view of culture, prior work has largely evaluated cultural commonsense within simplified notions of commonsense reasoning, typically using short, minimally contextualized tasks~\cite{sap-etal-2019-socialiqa, Bisk2020, mahendra-etal-2021-indonli, putri-etal-2024-llm, myung-etal-2024-blend, wibowo-etal-2024-copal, guo-etal-2025-care, permadi-et-al-2026-idmocqa}. For example,~\citet{wibowo-etal-2024-copal} relies on isolated prompts paired with single-turn multiple-choice questions. While useful for controlled benchmarking, such setups treat cultural knowledge as static, detached from the interactions through which it is expressed and negotiated. 
This mismatch becomes especially pronounced in culturally dense and multilingual environments, where practices vary across communities, as seen in Indonesia~\cite{koto-etal-2024-indoculture}.

Indonesia is a highly diverse nation, with over 700 languages, numerous ethnic groups, and distinct regionally grounded cultural norms~\cite{aji-etal-2022-one}. It is also the second most linguistically diverse country in the world~\cite{winata-etal-2023-nusax}. In such a setting, cultural commonsense is deeply localized, shaped by both national context and community-specific practices. Consequently, evaluating LLMs solely on Indonesian-language benchmarks
~\cite{ponti-etal-2020-xcopa, koto-etal-2022-indocloze, wibowo-etal-2024-copal, koto-etal-2024-indoculture} 
provides only a partial view of their cultural understanding. A more faithful assessment requires context-rich approaches that incorporate both Indonesian and local languages, capturing how cultural knowledge varies and unfolds across communities.

To address these challenges, we introduce \indoculturedialoguetitle, a new benchmark for culturally grounded dialogue in Indonesia, developed through a multi-stage human pipeline involving native speakers to ensure cultural authenticity. Compared with recent benchmarks~\cite{permadi-et-al-2026-idmocqa, aji-cohn-2025-loraxbench}, \indoculturedialoguetitle\ incorporates extensive human revision, annotation, and translation across the dataset creation process. While prior work either focuses on multihop reasoning in Indonesian only~\cite{permadi-et-al-2026-idmocqa} or lacks a conversational setting~\cite{aji-cohn-2025-loraxbench}, our benchmark emphasizes multi-turn dialogue, enabling more natural and context-rich cultural reasoning. It covers both general Indonesian culture and the cultures of 10 provinces, including Aceh, West Sumatra, West Java, Central Java, East Java, Bali, Nusa Tenggara Timur (NTT), South Kalimantan, South Sulawesi, and West Papua, spanning Indonesian and 10 local languages across 13 cultural topics.


As illustrated in Figure~\ref{fig:indoculturedial-overview}, each dialogue is constructed in both Indonesian and the corresponding local language spoken in the region (e.g., Wamesa in West Papua, a language with limited digital presence and considered vulnerable~\cite{ritchie-etal-2024-linguameta}). The final utterance in each dialogue is intentionally omitted, and the model is tasked with selecting the most appropriate continuation from multiple candidates.



We summarize our contributions as follows:
\begin{itemize}[noitemsep, topsep=0pt]
    \item We introduce \indoculturedialoguetitle, a dialogue-based benchmark for culturally grounded interactions in Indonesia, developed through a multi-stage human pipeline involving native speakers. It covers Indonesian and 10 local languages across 13 cultural topics.    
    \item We propose three evaluation tasks: (1) dialogue-based multiple-choice questions for cultural commonsense reasoning, (2) bidirectional machine translation between Indonesian and local languages, and (3) language steering with cultural grounding. These tasks assess LLMs’ ability to reason about culturally appropriate responses in dialogue, handle multilingual variation, and generate contextually grounded outputs.
    \item We evaluate a diverse set of LLMs, including proprietary, general multilingual, and Southeast Asian (SEA)-centric models. Our results show that open-source models lag behind, revealing a substantial gap in capturing and generating culturally grounded knowledge in Indonesian conversational settings.
\end{itemize}

\section{Related Works}

\paragraph{Dialogue Benchmarks in Multilingual Settings}
Most existing dialogue benchmarks focus on high-resource languages, particularly English (e.g., DailyDialog~\cite{li-etal-2017-dailydialog}, PersonaChat~\cite{zhang-etal-2018-personalizing-personachat}, and MultiWOZ~\cite{budzianowski-etal-2018-multiwoz}), and lack explicit cultural grounding. Recent efforts extend these benchmarks to multilingual settings, such as XDailyDialog~\cite{liu-etal-2023-xdailydialog} and mDIA~\cite{zhang2022mdiabenchmarkmultilingualdialogue-mdia}, which expands coverage to up to 46 languages. However, these resources provide limited support for Indonesian and its diverse local languages and cultures. As a result, they do not capture multi-turn dialogue, cultural commonsense reasoning, and localized linguistic variation in this context, which \indoculturedialoguetitle\ is designed to address.

\paragraph{Conversational-based NLP task in Indonesian Context}

Most Indonesian conversational datasets are developed as subsets of broader multilingual dialogue benchmarks. For instance, COD~\cite{majewska-etal-2023-cross-cod} is a multilingual task-oriented dialogue (TOD) benchmark covering four languages: Arabic, Indonesian, Russian, and Kiswahili. The benchmark includes tasks such as intent detection, slot labeling, and end-to-end dialogue modeling.~\citet{kautsar-etal-2023-indotod} focuses on task-oriented dialogue in a parallel Indonesian--English setting, where the system needs to understand the intent of the user and succeed in doing the user's task. More recently, \citet{kautsar2025seadialoguesmultilingualculturallygrounded} introduced a culturally grounded multilingual dialogue benchmark for Southeast Asian languages. However, the benchmark has limited language coverage and focuses on evaluating dialogue generation quality, with dialogues automatically constructed using LLMs. In contrast, \indoculturedialoguetitle\ emphasizes culturally grounded, multi-turn dialogue in the Indonesian context, combining dialogue-based cultural MCQ, machine translation, and language steering across Indonesian and local languages.



\paragraph{Commonsense Reasoning in Indonesian and Local Languages} 
Commonsense reasoning has been widely studied since the introduction of the Winograd Schema Challenge~\citep{levesque-2012-winograd}.
This line of work has since evolved, and such approaches have been extended to Indonesian through datasets such as XStoryCloze~\cite{lin-etal-2022-shot-xstorycloze}, IndoCloze~\cite{koto-etal-2022-indocloze}, XCOPA~\cite{ponti-etal-2020-xcopa}, and COPAL~\cite{wibowo-etal-2024-copal}. However, these datasets remain limited in capturing cultural nuances and localized aspects of Indonesian contexts.
\citet{koto-etal-2024-indoculture} introduce geographically grounded cultural commonsense reasoning across eleven Indonesian provinces, while \citet{permadi-et-al-2026-idmocqa} focus on multi-hop reasoning for Indonesian cultural understanding. However, these benchmarks rely on single-turn formats and do not incorporate dialogue-based interactions or coverage of Indonesian local languages. In parallel, \citet{aji-cohn-2025-loraxbench} introduce LoraxBench, a benchmark for evaluating cultural understanding in Indonesian and local languages through the translation of existing datasets. While expanding language coverage, it does not model conversational interactions or incorporate province-specific scenarios, limiting its ability to capture localized and context-dependent cultural communication. In contrast, \indoculturedialoguetitle\ explicitly incorporates multi-turn dialogue, as illustrated in Figure~\ref{fig:indoculturedial-overview}, to capture how cultural knowledge is expressed, contextualized, and negotiated in real-world interactions.

\section{Dataset Creation}

\begin{figure*}[t]
  \centering
  \includegraphics[width=1.9\columnwidth]{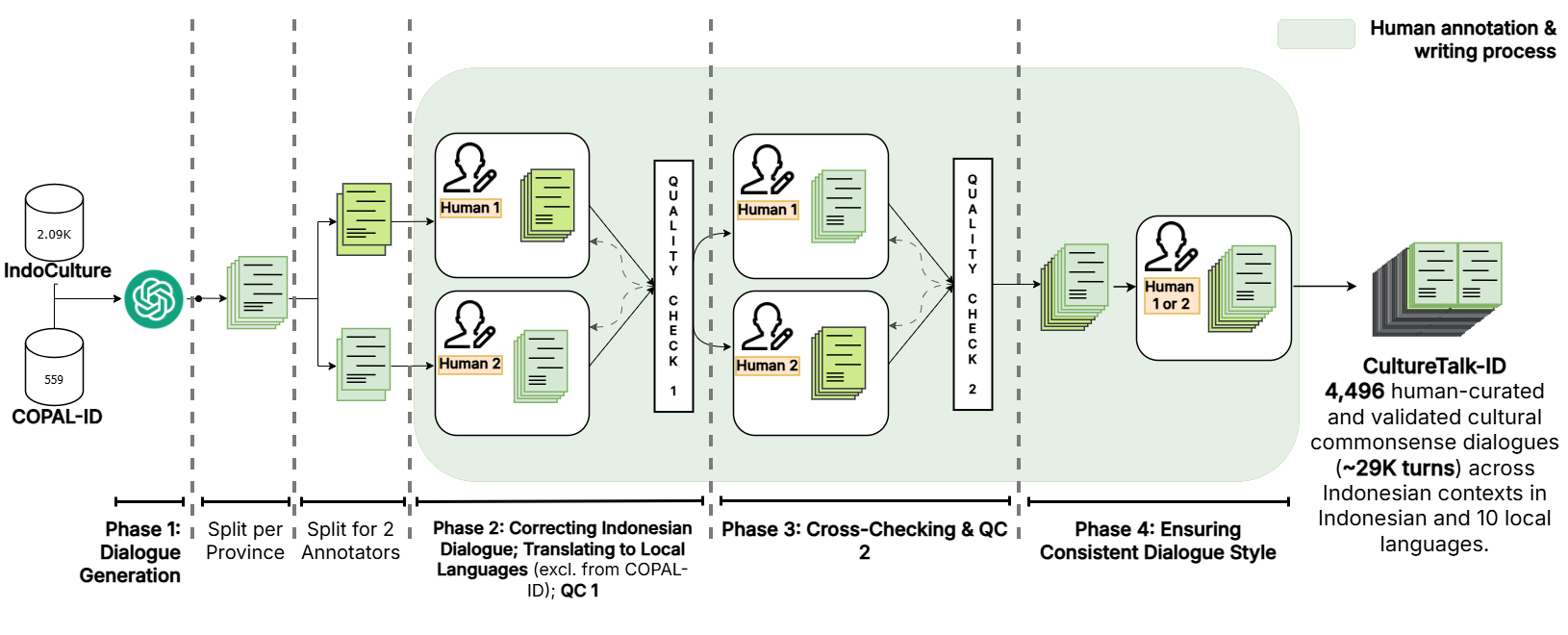}
  \caption{The \indoculturedialoguetitle\ dataset creation pipeline. It consists of four phases: dialogue generation; correction of Indonesian dialogues, translation into local languages, and initial quality control (QC1); cross-checking and secondary quality control (QC2); and finally, ensuring consistency in dialogue style.}
  \label{fig:indoculturedial-datacreation}
\end{figure*}

We construct the \indoculturedialoguetitle\ dataset using the pipeline outlined in Figure~\ref{fig:indoculturedial-datacreation}, transforming the IndoCulture~\cite{koto-etal-2024-indoculture} and COPAL-ID~\cite{wibowo-etal-2024-copal} datasets with the help of the human annotators into Indonesian and various Indonesian local languages. It covers 4,496 dialogues, 10 regions across Indonesia, 11 languages, and 13 topics, closely related to Indonesian culture.

\subsection{Phase 1: Dialogue Generation}

To initiate the pipeline, we select IndoCulture~\cite{koto-etal-2024-indoculture} and COPAL-ID~\cite{wibowo-etal-2024-copal} as pivot datasets, 
due to their culturally grounded multiple-choice formats.
We repurpose their premises as dialogue topics and treat the answer options as candidate utterances to facilitate dialogue construction and downstream human annotation. Given each instance, we prompt GPT-5 to generate a dialogue in Indonesian, intentionally omitting the final utterance. The original answer options are then adapted as candidate endings, with only one being contextually appropriate. 
We generate the dialogues in Indonesian because local Indonesian languages are generally low-resource and less reliably handled by current LLMs, making human involvement necessary for subsequent annotation and translation.
The resulting dialogues are grouped by province and prepared for human annotation and translation. The detailed process of this phase is provided in Appendix~\ref{sec:appendix-dialogue-generation}.

\subsection{Phase 2: Correction of Indonesian Dialogues, Translation to Local, and QC1}

The dialogues generated by GPT-5 were subsequently verified through human evaluation. For this process, we recruit two annotators for each cultural group, covering 10 provinces and one general Indonesian category (22 annotators in total). All annotators were required to be fluent in both Indonesian and their respective local languages, and to have lived in the corresponding province for more than 10 years.

During the recruitment process, candidates were provided with the task description and annotation guidelines, and were asked to complete a pilot annotation task to assess their understanding of the objectives. We then selected the most suitable annotators based on the quality of their pilot annotation outputs.
Then, all annotators participated in a preliminary briefing session to ensure a consistent understanding of the annotation guidelines. During the validation stage, annotators were instructed to verify cultural and answer correctness, and ensure that incorrect answer choices were culturally inappropriate or implausible.

Once the Indonesian dialogues had been validated, annotators translated them into their respective local languages. We allow flexible, non-literal translation to preserve fluency and naturalness in the target language, while keeping the same meaning in each utterance. After the annotators have completed the task, as a final quality-assurance step, we conduct additional sampling-based reviews of the annotated dialogues. Dialogues that did not satisfy the annotation standards were returned to annotators for revision.

\subsection{Phase 3: Cross-checking and QC2}

Following the first phase, we conduct a cross-check between annotators, followed by a second quality control stage (QC2). Since each language was assigned two annotators, each annotator was asked to review their partner's work for cultural correctness and translation quality.

To assess annotator attentiveness during this process, we deliberately inserted identifiable incorrect entries into approximately 5\% of the instances, following the methodology introduced by~\citet{Muszyski2023AttentionCA}. If an annotator failed to detect and correct these deliberately incorrect entries, they were asked to re-evaluate the entire set of dialogues until all injected errors had been identified and fixed. 
Additionally, we asked annotators to answer the MCQ instances themselves to verify that the correct answers were consistently identifiable. Instances answered incorrectly by the annotators were subsequently revised.
These processes served as our second-stage quality control mechanism. 
The transformation process from the original dataset until this phase is illustrated in Figure~\ref{fig:appendix-dialogue-generation-transformation} (Appendix~\ref{sec:appendix-dialogue-generation}).

\subsection{Phase 4: Consistency in Dialogue Style}

To ensure that the benchmark is not trivial, we identified two common issues arising from the LLM-generated dialogues in the early stage of our pipeline. The first issue is the non-uniform answer choices. In some cases, the correct option was written in a confident and natural manner, while the incorrect options contained obviously unnatural or doubtful expressions, making the correct answer easy to infer. 
For example, several incorrect options included phrases such as \textit{``malah''} (``instead/contrarily'') or \textit{``anehnya''} (``weirdly''), which implicitly signaled implausibility.
The second issue concerns the explicit presence of cultural entities within the answer choices. This may introduce shortcut cues within the utterance options, as illustrated in Appendix~\ref{sec:appendix-dialogue-generation}. 

To address these issues, we asked one annotator from each province to directly revise the answer options in both Indonesian and their corresponding local-language versions. The refined datasets were then merged into the final \indoculturedialoguetitle\ benchmark.

\begin{table}[t]
\centering
\small
\begin{tabular}{lc}
\toprule
\#dialogues & 4,496 \\
\#region-specific dialogue & 2,980 \\
\#turns & 29,420 \\
avg. words per dialogue & 60.23 \\
avg. utterance per dialogue & 6.54 \\
avg. words per utterance & 8.28 \\
\#words & 270,793 \\
\#unique words & 27,391 \\
\midrule
train : test split & 1:9 \\
\midrule
\#unique words & 9,385 (39.51\% of \\
(Indonesian $\wedge$ Local & local languages' \\
Indonesian languages) & unique words) \\
\bottomrule
\end{tabular}
\caption{Statistics of the \indoculturedialoguetitle\ dataset, combining Indonesian and local Indonesian language dialogues across all cultural regions.}
\label{tab:data-stats}
\end{table}

\subsection{\indoculturedialoguetitle\ Benchmark}

Each instance in \indoculturedialoguetitle\ consists of a culturally grounded dialogue with the final utterance omitted, together with three candidate responses, only one of which is culturally and contextually appropriate. Every dialogue is available in both Indonesian and its corresponding local Indonesian language, forming a parallel dialogue dataset spanning 10 Indonesian cultural regions. Table~\ref{tab:data-stats} presents the dataset statistics, where it contains 4,496 dialogues (2,980 region-specific dialogues), requiring knowledge tied to particular Indonesian cultural contexts. 
The more detailed statistics can be found in Appendix~\ref{sec:appendix-detailed-data-stats}.

To the best of our knowledge, \indoculturedialoguetitle\ is the first dialogue-based benchmark with extensive human curation for evaluating Indonesian cultural commonsense across both Indonesian and multiple local languages. In addition to dialogue-based cultural commonsense reasoning, its parallel structure also supports machine translation and language steering tasks.

\section{Experimental Setup}



Using \indoculturedialoguetitle, we evaluate dialogue-based cultural competence across Indonesian and local Indonesian languages. We consider three groups of models: (1) proprietary models, (2) multilingual LLMs, and (3) Southeast Asian (SEA)-centric LLMs. For proprietary models, we evaluate GPT-5.1, Gemini-2.5-flash, and Cohere-Command-A. The multilingual models include Qwen3-8B~\cite{qwen3technicalreport}, Llama-3.1-8B Instruct~\cite{grattafiori2024llama3herdmodels}, and Gemma-2-9B Instruct~\cite{gemma_2024}. Furthermore, the SEA-centric models include Sailor2-8B Chat~\cite{dou2025sailor2sailingsoutheastasia}, SEA-LION-v3.5-8B~\cite{SEA-LION-2504.05747}, and Sahabat-AI-v1 Instruct\footnote{\href{https://huggingface.co/Sahabat-AI/gemma2-9b-cpt-sahabatai-v1-instruct}{Sahabat-AI HuggingFace}}. All open-weight models are selected at a comparable parameter scale, around 8--9B parameters, while proprietary models are included to represent the current state of the art in large language models.

Building on the parallel dialogue dataset described in the previous section, we evaluate models on three complementary tasks. Prompts are provided in Appendix~\ref{sec:appendix-task-prompt}.

\setlength{\tabcolsep}{0.35em}
\begin{table*}[t]
\centering
\footnotesize
\begin{tabular}{l ccc ccc}
\toprule
\multirow{1}{*}{\yellowcell \textbf{Dialogue-based MCQ}}
& \multicolumn{3}{c}{\textbf{Context: None}}
& \multicolumn{3}{c}{\textbf{Context: Prov. + Lang.}} \\
\cmidrule(lr){2-4} \cmidrule(lr){5-7}
\textbf{Model} & \textbf{Avg} & \textbf{PS} & \textbf{\textasciitilde PS}
& \textbf{Avg} & \textbf{PS} & \textbf{\textasciitilde PS} \\
\midrule

\multicolumn{7}{c}{\textbf{Proprietary Models}} \\
GPT-5.1
& 82.31 \textsubscript{($\downarrow$6.92)} & 79.07 \textsubscript{($\downarrow$7.37)} & 88.67 \textsubscript{($\downarrow$6.05)}
& 82.70 \textsubscript{($\downarrow$7.25)} & 79.26 \textsubscript{($\downarrow$8.12)} & 89.45 \textsubscript{($\downarrow$5.54)} \\

Gemini-2.5-flash
& \textbf{87.77} \textsubscript{($\downarrow$3.69)} & \textbf{85.23} \textsubscript{($\downarrow$4.43)} & \textbf{92.74} \textsubscript{($\downarrow$2.25)}
& \textbf{87.72} \textsubscript{($\downarrow$3.96)} & \textbf{85.23} \textsubscript{($\downarrow$4.76)} & \textbf{92.61} \textsubscript{($\downarrow$2.38)} \\

Cohere-Command-A
& 75.09 \textsubscript{($\downarrow$9.65)} & 70.07 \textsubscript{($\downarrow$10.67)} & 84.96 \textsubscript{($\downarrow$7.65)}
& 74.64 \textsubscript{($\downarrow$10.37)} & 70.00 \textsubscript{($\downarrow$11.41)} & 83.77 \textsubscript{($\downarrow$8.31)} \\

\hline
\textit{average proprietary} 
& \textit{81.72} \textsubscript{($\downarrow$6.76)} & \textit{78.12} \textsubscript{($\downarrow$7.49)} & \textit{88.79} \textsubscript{($\downarrow$5.32)}
& \textit{81.69} \textsubscript{($\downarrow$7.19)} & \textit{78.16} \textsubscript{($\downarrow$8.10)} & \textit{88.61} \textsubscript{($\downarrow$5.41)}\\

\midrule

\multicolumn{7}{c}{\textbf{Multilingual Models}} \\
Qwen3-8B
& 52.89 \textsubscript{($\downarrow$16.06)} & 48.86 \textsubscript{($\downarrow$17.78)} & 60.82 \textsubscript{($\downarrow$13.02)}
& 54.72 \textsubscript{($\downarrow$15.56)} & 50.74 \textsubscript{($\downarrow$17.25)} & 62.53 \textsubscript{($\downarrow$12.27)} \\

Llama3.1-8B Instruct
& 51.87 \textsubscript{($\downarrow$10.76)} & 49.60 \textsubscript{($\downarrow$10.53)} & 56.33 \textsubscript{($\downarrow$11.22)}
& 53.87 \textsubscript{($\downarrow$11.57)} & 51.95 \textsubscript{($\downarrow$11.94)} & 57.65 \textsubscript{($\downarrow$11.09)} \\

Gemma2-9B Instruct
& \textbf{65.17} \textsubscript{($\downarrow$11.61)} & \textbf{59.80} \textsubscript{($\downarrow$13.02)} & \textbf{75.73} \textsubscript{($\downarrow$8.83)}
& \textbf{65.88} \textsubscript{($\downarrow$11.83)} & \textbf{60.87} \textsubscript{($\downarrow$13.22)} & \textbf{75.73} \textsubscript{($\downarrow$9.10)} \\

\hline
\textit{average multilingual}
& \textit{56.64} \textsubscript{($\downarrow$12.81)} & \textit{52.75} \textsubscript{($\downarrow$13.78)} & \textit{64.29} \textsubscript{($\downarrow$11.02)} 
& \textit{58.15} \textsubscript{($\downarrow$12.99)} & \textit{54.52} \textsubscript{($\downarrow$14.14)} & \textit{65.30} \textsubscript{($\downarrow$10.82)} \\

\midrule

\multicolumn{7}{c}{\textbf{SEA-Centric Models}} \\
Sailor2-8B Chat
& \textbf{70.15} \textsubscript{($\downarrow$9.92)} & \textbf{64.50} \textsubscript{($\downarrow$11.94)} & \textbf{81.27} \textsubscript{($\downarrow$5.93)}
& \textbf{72.02} \textsubscript{($\downarrow$9.30)} & \textbf{66.85} \textsubscript{($\downarrow$10.94)} & \textbf{82.19} \textsubscript{($\downarrow$6.07)} \\

SEA-LION-v3.5-8B
& 36.57 \textsubscript{(0.00)} & 36.85 \textsubscript{(0.00)} & 36.02 \textsubscript{(0.00)}
& 36.57 \textsubscript{(0.00)} & 36.85 \textsubscript{(0.00)} & 36.02 \textsubscript{(0.00)} \\

Sahabat-AI-v1 Instruct
& 69.08 \textsubscript{($\downarrow$11.79)} & 63.62 \textsubscript{($\downarrow$13.63)} & 79.82 \textsubscript{($\downarrow$8.17)}
& 69.93 \textsubscript{($\downarrow$11.61)} & 65.17 \textsubscript{($\downarrow$13.35)} & 79.29 \textsubscript{($\downarrow$8.18)} \\

\hline
\textit{average SEA-centric*}
& \textit{69.62} \textsubscript{($\downarrow$10.86)} & \textit{63.56} \textsubscript{($\downarrow$12.79)} & \textit{80.55} \textsubscript{($\downarrow$7.05)} 
& \textit{70.98} \textsubscript{($\downarrow$10.46)} & \textit{66.01} \textsubscript{($\downarrow$12.15)} & \textit{80.71} \textsubscript{($\downarrow$7.13)} \\

\bottomrule
\end{tabular}
\caption{\textbf{Dialogue-based MCQ} performance on local Indonesian language dialogues. \textsubscript{Values in parentheses} denote the performance difference compared to Indonesian dialogue settings, where $\downarrow$ indicates performance degradation in local language settings. \textbf{Bold} scores indicate the best score within each model category. *The SEA-LION-v3.5-8B-R scores are excluded.}
\label{tab:main-local-dialogue-delta}
\end{table*}

\paragraph{MCQ Evaluation.}
For the MCQ evaluation task, each model is given a dialogue context with the final utterance omitted, together with three candidate responses, only one of which is culturally and contextually appropriate. For open-source models, rather than using generation-based evaluation, we adopt a likelihood-based setup in which models score each candidate response and select the option with the highest likelihood as the prediction.
We evaluate both in Indonesian and local-language dialogues, and 
we report average accuracy as the main evaluation metric.

\paragraph{Machine Translation.}
The machine translation task evaluates a model's ability to translate dialogue utterances between Indonesian and local Indonesian languages (as listed in Appendix~\ref{sec:appendix-language-covered-indoculturedialogue}) in both directions. 
Translation quality is evaluated using BLEU~\cite{papineni-etal-2002-bleu}, BERTScore~\cite{zhang-etal-2020-bertscore}, and an LLM-as-a-judge framework 
on a 1-5 scale. We evaluate models in zero-shot settings and additionally report supervised fine-tuning (SFT) results for selected open-weight models. For open-weight models, inference is repeated three times, and the mean and standard deviation are reported.

\paragraph{Language Steering.}
The language steering task evaluates whether models can generate responses in a requested linguistic variety. This is important, as Indonesians are bilingual, able to speak in Indonesian and their own local language. Given a dialogue context, models are instructed to produce the next utterance either in Indonesian or in a specified local Indonesian language. 
We evaluate generated responses using an LLM-as-a-judge score, Glot-LID~\citep{kargaran-etal-2023-glotlid} language identification accuracy, BLEU, and chrF. 

\setlength{\tabcolsep}{0.35em}
\begin{table*}[t]
\centering
\footnotesize
\begin{tabular}{l c c ccccc}
\toprule
\multirow{1}{*}{\purplecell \textbf{Machine Translation}}
& \textbf{BLEU Scores} 
& \textbf{BERTScore} 
& \multicolumn{5}{c}{\textbf{LLM as a Judge (1--5)}} \\
\cmidrule(lr){2-2} \cmidrule(lr){3-3} \cmidrule(lr){4-8}
Model & \textbf{B4} & \textbf{F1}
& \textbf{adequacy} & \textbf{fluency} & \textbf{register} & \textbf{terminology} & \textbf{overall} \\
\midrule
\midrule
\multicolumn{8}{l}{\textit{Indonesian $\rightarrow$ local Indonesian languages}} \\
\midrule
\multicolumn{8}{c}{\textbf{Proprietary Models (0-shot)}} \\
GPT-5.1 
& 27.53 & 84.02 & 4.35 & 3.80 & 3.75 & 4.33 & 3.87 \textsubscript{(2)} \\
Gemini-2.5-flash 
& \textbf{28.36} & \textbf{84.33} & \textbf{4.48} & \textbf{3.91} & \textbf{3.91} & \textbf{4.48} & \textbf{4.00} \textsubscript{(1)} \\
Cohere-Command-A 
& 21.45 & 81.40 & 3.35 & 2.82 & 2.69 & 3.29 & 2.93 \textsubscript{(5)} \\
\midrule

\multicolumn{8}{c}{\textbf{Multilingual Models (0-shot)}} \\
Qwen3-8B 
& 19.78 \textsubscript{($\pm$0.1)} & 78.74 \textsubscript{($\pm$0.1)} & 2.82 & 2.11 & 1.90 & 2.73 & 2.22 \textsubscript{(11)} \\
Llama-3.1-8B Instruct 
& 17.62 \textsubscript{($\pm$0.1)} & 78.71 \textsubscript{($\pm$0.1)} & 2.79 & 2.27 & 2.21 & 2.81 & 2.49 \textsubscript{(10)} \\
Gemma-2-9b Instruct 
& \textbf{20.54} \textsubscript{($\pm$0.1)} & \textbf{80.15} \textsubscript{($\pm$0.1)} & \textbf{3.20} & \textbf{2.49} & \textbf{2.40} & \textbf{3.11} & \textbf{2.73} \textsubscript{(7)} \\
\midrule

\multicolumn{8}{c}{\textbf{SEA-Centric Models (0-shot)}} \\
Sailor2-8B Chat 
& 15.44 \textsubscript{($\pm$0.1)} & 77.58 \textsubscript{($\pm$0.1)} & 3.38 & \textbf{2.85} & \textbf{2.65} & \textbf{3.42} & \textbf{2.94} \textsubscript{(4)} \\
SEA-LION-v3.5-8B 
& 19.34 \textsubscript{($\pm$0.1)} & 78.98 \textsubscript{($\pm$0.1)} & 3.18 & 2.61 & 2.44 & 3.11 & 2.72 \textsubscript{(9)} \\
Sahabat-AI-v1 Instruct 
& \textbf{21.70} \textsubscript{($\pm$0.0)} & \textbf{80.51} \textsubscript{($\pm$0.0)} & \textbf{3.57} & 2.63 & 2.46 & 3.21 & 2.80 \textsubscript{(6)} \\
\midrule

\multicolumn{8}{c}{\textbf{SFT Models}} \\
Gemma-2-9b Instruct 
& \green{$\Uparrow$} 22.60 \textsubscript{($\pm$0.0)} & \red{$\Downarrow$} 79.39 \textsubscript{($\pm$0.0)} & \red{$\Downarrow$} 3.16 & \green{$\Uparrow$} 2.54 & \green{$\Uparrow$} 2.46 & \green{$\Uparrow$} 3.14 & \textcolor{gray}{$\approx$} 2.73 \textsubscript{(7)} \\
Sahabat-AI-v1 Instruct 
& \green{$\Uparrow$} \textbf{24.56} \textsubscript{($\pm$0.0)} & \red{$\Downarrow$} \textbf{80.30} \textsubscript{($\pm$0.0)} & \red{$\Downarrow$} \textbf{3.53} & \green{$\Uparrow$} \textbf{3.03} & \green{$\Uparrow$} \textbf{2.97} & \green{$\Uparrow$} \textbf{3.55} & \green{$\Uparrow$} \textbf{3.14} \textsubscript{(3)} \\
\bottomrule

\end{tabular}
\caption{\textbf{Machine translation} results for Indonesian $\rightarrow$ local languages under the Province + Language context. We report BLEU-4, BERTScore F1, and LLM-as-a-judge scores for adequacy, fluency, register, terminology, and overall quality. The \textsubscript{scores} in BLEU and BERTScore represent standard deviation, while the \textsubscript {numbers} in the overall score represent the ranking of each model. \textbf{Bold} scores indicate the best score within each model category.}
\label{tab:mt-main-result}
\end{table*}

\section{Results \& Analysis}
\label{sec:results}

\subsection{Dialogue-based Cultural Commonsense MCQ Evaluation}

The evaluation results for dialogue-based cultural commonsense are presented in Table~\ref{tab:main-local-dialogue-delta}. Overall, proprietary models achieve the best performance on this benchmark, delivering consistent results across both settings, with Gemini-2.5-flash performing the best among the proprietary models. However, generic multilingual models lag substantially behind proprietary models, where in contrast, SEA-centric models consistently outperform generic multilingual baselines, with Sailor2-8B-Chat achieving the strongest performance on local Indonesian dialogues.
However, SEA-LION-v3.5-8B-R exhibits degenerate behavior, consistently selecting the same answer choice (B) across questions.

We also find that performance drops substantially in local Indonesian language settings. 
Under the no-context setting (i.e., without province and language information), performance decreases by 6.76, 12.81, and 10.86 points for proprietary, generic multilingual, and SEA-centric models, respectively. This underscores the challenges of understanding low-resource languages, particularly Indonesian local languages, despite their large number of speakers.

We also observe that providing additional context improves performance for open-source model groups, particularly multilingual and SEA-centric models, while the gains are relatively limited for proprietary models. 
Province-specific (PS) dialogues benefit the most from this additional context. Moreover, as more contextual information is introduced, the performance gap between PS and \textasciitilde PS dialogues narrows, suggesting that additional context helps models better cultural grounding and capture localized cultural nuances. The more detailed analysis can be found in Appendix~\ref{sec:appendix-mcq-evaluation-details-a}.

\paragraph{Is further pretraining an LLM in a cultural context helpful?}
We investigate whether additional pretraining in an Indonesian cultural context improves model performance, and to what extent it benefits downstream tasks. To this end, we compare Gemma-2-9B Instruct with Sahabat-AI-v1 Instruct, where the latter is further adapted from the former using Indonesian-centric training data. As shown in Figure~\ref{fig:mcq-multi-vs-sea} (Appendix~\ref{sec:appendix-mcq-evaluation-details-b}), performance improves consistently across all provinces, 
where somehow the largest improvement is in South Kalimantan, despite the additional pretraining primarily introducing only Indonesian, Javanese, and Sundanese contexts.
Overall, these results indicate that culturally grounded pretraining provides consistent and transferable benefits, supporting the conclusion that such pretraining is indeed effective.

\subsection{Machine Translation Evaluation}

We present machine translation results in Table~\ref{tab:mt-main-result} for Indonesian to local Indonesian languages\footnote{We conduct statistical significance testing for all metrics with available standard deviations.}, with even the best model achieving moderate performance, while most models fall below 3.5. Proprietary models, such as Gemini-2.5-flash and GPT-5.1, achieve the strongest results across evaluation metrics. Generic multilingual models lag significantly, while SEA-centric models perform better, which narrows the gap with proprietary systems. Across evaluation dimensions, fluency and register consistently receive lower scores than adequacy and terminology, suggesting that generating natural and contextually appropriate local expressions remains a key challenge.

Supervised fine-tuning on the \indoculturedialoguetitle\ training set yields consistent improvements. Both Gemma-2-9B Instruct and Sahabat-AI-v1 Instruct outperform their base versions. Additionally, Sahabat-AI-v1 Instruct also surpasses Cohere-Command-A in this setting. Substantial gains are also observed in fluency and register, indicating that supervised fine-tuning, even with a relatively small train set (448 dialogues), effectively enhances naturalness and stylistic appropriateness. The machine translation experiment details further can be found in Appendix~\ref{sec:appendix-mt-evaluation-details}.

\paragraph{Correlation between the LLM-as-a-judge and human evaluation.}
Table~\ref{tab:mt-main-result} shows a general consistency between computation-based metrics and LLM-as-a-judge scores. To further assess the reliability of LLM-as-a-judge in culturally grounded translation tasks, we analyze its correlation with human judgments. Specifically, we ask fluent Javanese and Minangkabau speakers to evaluate the adequacy and fluency of 60 generated samples from the best-performing models in each category. As shown in Appendix~\ref{sec:appendix-llm-correlation}, the mean absolute distance (MAD) is generally below 1, and the accuracy-at-one (Acc@1) average reaches 0.83, indicating that LLM-as-a-judge is a reliable proxy in this setting. We also observe that stronger models, typically proprietary models, exhibit closer alignment with human judgments.


\subsection{Language Steering Evaluation}

\setlength{\tabcolsep}{0.05em}
\begin{table}[t]
\centering
\footnotesize
\begin{tabular}{l cccc}
\toprule
\multirow{1}{*}{\blueell \textbf{Language Steering}}
& \multicolumn{3}{c}{\textbf{Indonesian $\rightarrow$ Local languages}} \\
\cmidrule(lr){2-4}
Model & \textbf{LLM} & \textbf{Glot-LID} & \textbf{BLEU}  \\
\midrule

\multicolumn{4}{c}{\textbf{Proprietary Models}} \\
GPT-5.1 
& 4.42 & 73.9 & 0.64 \\
Gemini-3-flash 
& \textbf{4.55} & \textbf{92.7} & \textbf{0.94} \\
Cohere-Command-A 
& 3.70 & 31.6 & 0.62 \\
\midrule

\multicolumn{4}{c}{\textbf{Multilingual Models}} \\
Qwen3-8B 
& \textbf{3.11} \textsubscript{($\pm$0.0)} & 1.5 \textsubscript{($\pm$0.2)} & 0.42 \textsubscript{($\pm$0.1)} \\
Llama-3.1-8B Inst.
& 2.91 \textsubscript{($\pm$0.0)} & \textbf{9.2} \textsubscript{($\pm$0.4)} & 0.32 \textsubscript{($\pm$0.2)} \\
Gemma-2-9b Inst. 
& 3.00 \textsubscript{($\pm$0.0)} & 7.8 \textsubscript{($\pm$0.6)} & \textbf{0.44} \textsubscript{($\pm$0.0)} \\
\midrule

\multicolumn{4}{c}{\textbf{SEA-Centric Models}} \\
Sailor2-8B 
& 3.19 \textsubscript{($\pm$0.0)} & 23.7 \textsubscript{($\pm$0.4)} & 0.23 \textsubscript{($\pm$0.0)} \\
SEA-LION-v3.5-8B 
& 3.01 \textsubscript{($\pm$0.0)} & \textbf{27.5} \textsubscript{($\pm$0.7)} & 0.26 \textsubscript{($\pm$0.0)} \\
Sahabat-AI-v1 Inst. 
& \textbf{3.32} \textsubscript{($\pm$0.0)} & 13.7 \textsubscript{($\pm$1.0)} & \textbf{0.64} \textsubscript{($\pm$0.2)} \\
\midrule

\multicolumn{4}{c}{\textbf{SFT Models}} \\
Qwen3-8B 
& \red{$\Downarrow$} 3.06 \textsubscript{($\pm$0.1)} & \green{$\Uparrow$} 4.4 \textsubscript{($\pm$5.1)} & \green{$\Uparrow$} 0.50 \textsubscript{($\pm$0.1)} \\
Sahabat-AI-v1 Inst. 
& \green{$\Uparrow$} \textbf{3.33} \textsubscript{($\pm$0.1)} & \green{$\Uparrow$} \textbf{18.2} \textsubscript{($\pm$8.9)} & \green{$\Uparrow$} \textbf{0.77} \textsubscript{($\pm$0.0)} \\
\bottomrule
\end{tabular}
\caption{\textbf{Language steering} to local languages results. We report LLM-as-a-judge (denoted as LLM), Glot-LID, and BLEU scores. The \textsubscript{scores} represent the standard deviation. \textbf{Bold} scores indicate the best score within each model category.}
\label{tab:main-dialogue-steering-local}
\end{table}

We present the language steering results from Indonesian to local Indonesian languages in Table~\ref{tab:main-dialogue-steering-local}. Overall, model performance drops substantially when generating local languages, in sharp contrast to the considerably stronger performance observed when generating Indonesian responses (see Appendix~\ref{sec:appendix-dialogue-steering-evaluation-details}). Only Gemini-3-flash-preview and GPT-5.1 achieve moderate success, while Cohere-Command-A frequently fails to produce outputs in the correct language (31.6\%), even before considering the correctness of the utterance content itself. Open-source models struggle considerably. The best-performing open-source model, SEA-LION-v3.5-8B, generates only 27.5\% of utterances in the correct local language, while Qwen3-8B achieves just 1.5\%. Interestingly, supervised fine-tuned models tend to perform better when generating local-language responses, although this improvement is often accompanied by a slight degradation in their generation semantic quality.

Upon further inspection, we find that many open-source models default to Indonesian, even when a local language is required. As illustrated in Table~\ref{tab:steering-example}, both Qwen3-8B and Sahabat-AI-v1 Instruct (despite being trained on Javanese) mostly produce output in Indonesian rather than Javanese. Moreover, the generated content is incorrect: the context requires adding more soup to prevent the food from drying out, but these models instead suggest adding \textit{``santan''} (coconut milk) or even more sambal, which is not appropriate for the given situation. We present the more detailed language steering analysis in Appendix~\ref{sec:appendix-dialogue-steering-evaluation-details}.

\begin{table}[t]
\centering
\small
\begin{tabular}{p{1.2cm} p{6cm}}
\toprule
\textbf{Input} & \dots \\
& Dewi: ``Nasi jagung kan agak kering ya, biasanya kamu gimana biar makannya enak?'' \\
& \textit{(Dewi: ``Corn rice can be a bit dry. What do you usually do to make it more enjoyable to eat?'')} \\
& Syavira: \dots \\
\midrule
\textbf{Gold} & ``Aku mangan sego jagung diseloni kuah ben ora keselek.'' \\
& \textit{from Javanese: (``I eat corn rice with soup so that I will not choke.'')} \\
\midrule
\textbf{Gemini-3-flash-preview} & ``Iyo bener, mangkane biasane tak tambahi jangan lodeh sing akeh kuahe karo iwak asin plus sambel bajak, wis jan mantep pol!'' \green{\ding{51}} \\
& \textit{from Javanese: (``Correct, that is why I usually add a lot of lodeh soup with salted fish and \textit{sambel bajak}, it is really good!'')} \\
\midrule
\textbf{Qwen3-8B} & ``Aku biasa rebus sedikit santan biar lembut dan lebih enak.'' \red{\ding{55}} \\
& \textit{from Indonesian: (``I usually boil a bit of coconut milk to make it softer and tastier.'')} \\
\midrule
\textbf{Sahabat-AI-v1} & ``Aku biasanya tambahin sambal biar pedes, Dewi.'' \red{\ding{55}} \\
\textbf{Instruct} & \textit{from Indonesian: (``I usually add sambal to make it spicy, Dewi.'')} \\
\bottomrule
\end{tabular}
\caption{Example of language steering for Javanese. Open-source models tend to default to Indonesian and produce contextually incorrect responses.}
\label{tab:steering-example}
\end{table}

\section{Conclusion}

In this work, we present \indoculturedialoguetitle, a human-curated, culturally grounded dialogue dataset covering Indonesian and 10 province-specific cultural contexts. The dataset consists of 4,496 dialogues and is designed to benchmark language models' ability to understand and generate culturally nuanced interactions across regions. 
We ensure data quality through a multi-stage pipeline and incorporate human annotators.
We benchmark a range of proprietary and open-source models on three tasks: dialogue-MCQ cultural commonsense, machine translation, and language steering.

Our results show that while proprietary models consistently achieve strong performance, open-source models still struggle, particularly in generation tasks such as machine translation and language steering. These results highlight a gap in culturally grounded understanding and generation capabilities in dialogue, 
suggesting the need for future work in incorporating richer Indonesian cultural context and broader local language coverage into model pretraining and adaptation.

\section*{Limitations}
\label{sec:Limitations}



While our dataset is the first to cover cultural contexts in Indonesian and local Indonesian language dialogues, it still does not represent all provinces in Indonesia. Indonesia consists of 38 provinces with highly diverse cultural backgrounds, and several regions and cultural variations remain underrepresented in our data. In addition, the topic distribution is not entirely balanced. Some domains, such as `agriculture', `farm', and `game', are less represented compared to other topics, although the imbalance is relatively moderate. The difficulty level of dialogues across provinces may also be imbalanced, as some provinces may require less complex cultural reasoning to answer correctly, which could influence the resulting scores. Future work can expand both the topical coverage and the regional diversity represented in the benchmark.


For the human evaluation correlation in the MT experiments, we were only able to conduct detailed correlation analyses between LLM-as-a-judge and the human evaluation for Javanese and Minangkabau. Although our dataset creation effort involved 22 annotators from 11 provinces, recruiting qualified annotators for many local languages remains challenging. This is partly due to the differences in local language standardization, as discussed in~\cite{novitasari-etal-2020-cross}, and the speaker availability. Nevertheless, the results for Javanese and Minangkabau show reasonably strong correlations, particularly for the proprietary models.

\section*{Ethical Consideration}


This dataset is released under the CC BY-NC-SA 4.0 license. The dataset is intended for benchmarking large language model performance on Indonesian cultural understanding in dialogue contexts. We allow the dataset to be used as training data, provided that only the designated training split is used and that the usage is strictly non-commercial. Users should also be aware that models trained on this dataset may inherit biases toward the languages and cultural groups represented in the data, as our benchmark does not cover the full diversity of Indonesian cultures, as discussed in the Limitations section.

Additionally, all 22 annotators were compensated based on agreements established before the annotation process. The compensation was considered appropriate within the Indonesian economic context and proportional to the workload involved. The annotators were informed that the dataset would be publicly released. Furthermore, the dataset does not contain personally identifiable information, as the annotations were conducted on third-person point-of-view conversations rather than personal user interactions.

\bibliography{custom}

@article{koto-etal-2024-indoculture,
    title = "{I}ndo{C}ulture: Exploring Geographically Influenced Cultural Commonsense Reasoning Across Eleven {I}ndonesian Provinces",
    author = "Koto, Fajri  and
      Mahendra, Rahmad  and
      Aisyah, Nurul  and
      Baldwin, Timothy",
    journal = "Transactions of the Association for Computational Linguistics",
    volume = "12",
    year = "2024",
    address = "Cambridge, MA",
    publisher = "MIT Press",
    url = "https://aclanthology.org/2024.tacl-1.92/",
    doi = "10.1162/tacl_a_00726",
    pages = "1703--1719"
}

@inproceedings{wibowo-etal-2024-copal,
    title = "{COPAL}-{ID}: {I}ndonesian Language Reasoning with Local Culture and Nuances",
    author = "Wibowo, Haryo  and
      Fuadi, Erland  and
      Nityasya, Made  and
      Prasojo, Radityo Eko  and
      Aji, Alham",
    editor = "Duh, Kevin  and
      Gomez, Helena  and
      Bethard, Steven",
    booktitle = "Proceedings of the 2024 Conference of the North American Chapter of the Association for Computational Linguistics: Human Language Technologies (Volume 1: Long Papers)",
    month = jun,
    year = "2024",
    address = "Mexico City, Mexico",
    publisher = "Association for Computational Linguistics",
    url = "https://aclanthology.org/2024.naacl-long.77/",
    doi = "10.18653/v1/2024.naacl-long.77",
    pages = "1404--1422"
}

@article{nguyen-2022-bayesian-mindsponge,
title = {Introduction to Bayesian Mindsponge Framework analytics: An innovative method for social and psychological research},
journal = {MethodsX},
volume = {9},
pages = {101808},
year = {2022},
issn = {2215-0161},
doi = {https://doi.org/10.1016/j.mex.2022.101808},
url = {https://www.sciencedirect.com/science/article/pii/S2215016122001881},
author = {Minh-Hoang Nguyen and Viet-Phuong La and Tam-Tri Le and Quan-Hoang Vuong}
}

@article{liu-etal-2025-culturally,
    author = {Liu, Chen Cecilia and Gurevych, Iryna and Korhonen, Anna},
    title = {Culturally Aware and Adapted NLP: A Taxonomy and a Survey of the State of the Art},
    journal = {Transactions of the Association for Computational Linguistics},
    volume = {13},
    pages = {652-689},
    year = {2025},
    month = {07},
    issn = {2307-387X},
    doi = {10.1162/tacl_a_00760},
    url = {https://doi.org/10.1162/tacl_a_00760},
    eprint = {https://direct.mit.edu/tacl/article-pdf/doi/10.1162/tacl_a_00760/2535104/tacl_a_00760.pdf},
}

@incollection{suraj-2024-culture,
    author = {Sharma, Suraj and Liu, Leigh Anne},
    editor = {Gelfand, Michele J. and Erez, Miriam},
    isbn = {9780190085384},
    title = {Culture and Communication},
    booktitle = {The Oxford Handbook of Cross-Cultural Organizational Behavior},
    publisher = {Oxford University Press},
    year = {2024},
    month = {01},
    doi = {10.1093/oxfordhb/9780190085384.013.16},
    url = {https://doi.org/10.1093/oxfordhb/9780190085384.013.16},
    eprint = {https://academic.oup.com/book/0/chapter/436603421/chapter-ag-pdf/56531499/book_55822_section_436603421.ag.pdf},
}

@inproceedings{wu-etal-2025-socialcc,
    title = "{S}ocial{CC}: Interactive Evaluation for Cultural Competence in Language Agents",
    author = "Wu, Jincenzi  and
      Lian, Jianxun  and
      Wang, Dingdong  and
      Meng, Helen M.",
    editor = "Che, Wanxiang  and
      Nabende, Joyce  and
      Shutova, Ekaterina  and
      Pilehvar, Mohammad Taher",
    booktitle = "Proceedings of the 63rd Annual Meeting of the Association for Computational Linguistics (Volume 1: Long Papers)",
    month = jul,
    year = "2025",
    address = "Vienna, Austria",
    publisher = "Association for Computational Linguistics",
    url = "https://aclanthology.org/2025.acl-long.1594/",
    doi = "10.18653/v1/2025.acl-long.1594",
    pages = "33242--33271",
    ISBN = "979-8-89176-251-0"
}

@inproceedings{sap-etal-2019-socialiqa,
    title = "Social {IQ}a: Commonsense Reasoning about Social Interactions",
    author = "Sap, Maarten  and
      Rashkin, Hannah  and
      Chen, Derek  and
      Le Bras, Ronan  and
      Choi, Yejin",
    editor = "Inui, Kentaro  and
      Jiang, Jing  and
      Ng, Vincent  and
      Wan, Xiaojun",
    booktitle = "Proceedings of the 2019 Conference on Empirical Methods in Natural Language Processing and the 9th International Joint Conference on Natural Language Processing (EMNLP-IJCNLP)",
    month = nov,
    year = "2019",
    address = "Hong Kong, China",
    publisher = "Association for Computational Linguistics",
    url = "https://aclanthology.org/D19-1454/",
    doi = "10.18653/v1/D19-1454",
    pages = "4463--4473"
}

@inproceedings{Bisk2020,
  author = {Yonatan Bisk and Rowan Zellers and
            Ronan Le Bras and Jianfeng Gao
            and Yejin Choi},
  title = {PIQA: Reasoning about Physical Commonsense in
           Natural Language},
  booktitle = {Proceedings of the AAAI Conference on Artificial Intelligence},
  year = {2020},
  pages = {7432--7439},
  volume = {34},
  doi = {10.1609/aaai.v34i05.6239}
}

@inproceedings{mahendra-etal-2021-indonli,
    title = "{I}ndo{NLI}: A Natural Language Inference Dataset for {I}ndonesian",
    author = "Mahendra, Rahmad  and
      Aji, Alham Fikri  and
      Louvan, Samuel  and
      Rahman, Fahrurrozi  and
      Vania, Clara",
    editor = "Moens, Marie-Francine  and
      Huang, Xuanjing  and
      Specia, Lucia  and
      Yih, Scott Wen-tau",
    booktitle = "Proceedings of the 2021 Conference on Empirical Methods in Natural Language Processing",
    month = nov,
    year = "2021",
    address = "Online and Punta Cana, Dominican Republic",
    publisher = "Association for Computational Linguistics",
    url = "https://aclanthology.org/2021.emnlp-main.821/",
    doi = "10.18653/v1/2021.emnlp-main.821",
    pages = "10511--10527"
}

@inproceedings{putri-etal-2024-llm,
    title = "Can {LLM} Generate Culturally Relevant Commonsense {QA} Data? Case Study in {I}ndonesian and {S}undanese",
    author = "Putri, Rifki Afina  and
      Haznitrama, Faiz Ghifari  and
      Adhista, Dea  and
      Oh, Alice",
    editor = "Al-Onaizan, Yaser  and
      Bansal, Mohit  and
      Chen, Yun-Nung",
    booktitle = "Proceedings of the 2024 Conference on Empirical Methods in Natural Language Processing",
    month = nov,
    year = "2024",
    address = "Miami, Florida, USA",
    publisher = "Association for Computational Linguistics",
    url = "https://aclanthology.org/2024.emnlp-main.1145/",
    doi = "10.18653/v1/2024.emnlp-main.1145",
    pages = "20571--20590"
}

@inproceedings{myung-etal-2024-blend,
 author = {Myung, Junho and Lee, Nayeon and Zhou, Yi and Jin, Jiho and Putri, Rifki Afina and Antypas, Dimosthenis and Borkakoty, Hsuvas and Kim, Eunsu and Perez-Almendros, Carla and Ayele, Abinew Ali and Guti\'{e}rrez-Basulto, V\'{\i}ctor and Ib\'{a}\~{n}ez-Garc\'{\i}a, Yazm\'{\i}n and Lee, Hwaran and Muhammad, Shamsuddeen Hassan and Park, Kiwoong and Rzayev, Anar Sabuhi and White, Nina and Yimam, Seid Muhie and Pilehvar, Mohammad Taher and Ousidhoum, Nedjma and Camacho-Collados, Jose and Oh, Alice},
 booktitle = {Advances in Neural Information Processing Systems},
 doi = {10.52202/079017-2483},
 editor = {A. Globerson and L. Mackey and D. Belgrave and A. Fan and U. Paquet and J. Tomczak and C. Zhang},
 pages = {78104--78146},
 publisher = {Curran Associates, Inc.},
 title = {BLEnD: A Benchmark for LLMs on Everyday Knowledge in Diverse Cultures and Languages},
 url = {https://proceedings.neurips.cc/paper_files/paper/2024/file/8eb88844dafefa92a26aaec9f3acad93-Paper-Datasets_and_Benchmarks_Track.pdf},
 volume = {37},
 year = {2024}
}

@inproceedings{guo-etal-2025-care,
    title = "{CARE}: Multilingual Human Preference Learning for Cultural Awareness",
    author = "Guo, Geyang  and
      Naous, Tarek  and
      Wakaki, Hiromi  and
      Nishimura, Yukiko  and
      Mitsufuji, Yuki  and
      Ritter, Alan  and
      Xu, Wei",
    editor = "Christodoulopoulos, Christos  and
      Chakraborty, Tanmoy  and
      Rose, Carolyn  and
      Peng, Violet",
    booktitle = "Proceedings of the 2025 Conference on Empirical Methods in Natural Language Processing",
    month = nov,
    year = "2025",
    address = "Suzhou, China",
    publisher = "Association for Computational Linguistics",
    url = "https://aclanthology.org/2025.emnlp-main.1669/",
    doi = "10.18653/v1/2025.emnlp-main.1669",
    pages = "32866--32895",
    ISBN = "979-8-89176-332-6"
}

@inproceedings{aji-etal-2022-one,
    title = "One Country, 700+ Languages: {NLP} Challenges for Underrepresented Languages and Dialects in {I}ndonesia",
    author = "Aji, Alham Fikri  and
      Winata, Genta Indra  and
      Koto, Fajri  and
      Cahyawijaya, Samuel  and
      Romadhony, Ade  and
      Mahendra, Rahmad  and
      Kurniawan, Kemal  and
      Moeljadi, David  and
      Prasojo, Radityo Eko  and
      Baldwin, Timothy  and
      Lau, Jey Han  and
      Ruder, Sebastian",
    editor = "Muresan, Smaranda  and
      Nakov, Preslav  and
      Villavicencio, Aline",
    booktitle = "Proceedings of the 60th Annual Meeting of the Association for Computational Linguistics (Volume 1: Long Papers)",
    month = may,
    year = "2022",
    address = "Dublin, Ireland",
    publisher = "Association for Computational Linguistics",
    url = "https://aclanthology.org/2022.acl-long.500/",
    doi = "10.18653/v1/2022.acl-long.500",
    pages = "7226--7249"
}

@inproceedings{winata-etal-2023-nusax,
    title = "{N}usa{X}: Multilingual Parallel Sentiment Dataset for 10 {I}ndonesian Local Languages",
    author = "Winata, Genta Indra  and
      Aji, Alham Fikri  and
      Cahyawijaya, Samuel  and
      Mahendra, Rahmad  and
      Koto, Fajri  and
      Romadhony, Ade  and
      Kurniawan, Kemal  and
      Moeljadi, David  and
      Prasojo, Radityo Eko  and
      Fung, Pascale  and
      Baldwin, Timothy  and
      Lau, Jey Han  and
      Sennrich, Rico  and
      Ruder, Sebastian",
    editor = "Vlachos, Andreas  and
      Augenstein, Isabelle",
    booktitle = "Proceedings of the 17th Conference of the European Chapter of the Association for Computational Linguistics",
    month = may,
    year = "2023",
    address = "Dubrovnik, Croatia",
    publisher = "Association for Computational Linguistics",
    url = "https://aclanthology.org/2023.eacl-main.57/",
    doi = "10.18653/v1/2023.eacl-main.57",
    pages = "815--834"
}

@inproceedings{ritchie-etal-2024-linguameta,
    title = "{L}ingua{M}eta: Unified Metadata for Thousands of Languages",
    author = "Ritchie, Sandy  and
      van Esch, Daan  and
      Okonkwo, Uche  and
      Vashishth, Shikhar  and
      Drummond, Emily",
    editor = "Calzolari, Nicoletta  and
      Kan, Min-Yen  and
      Hoste, Veronique  and
      Lenci, Alessandro  and
      Sakti, Sakriani  and
      Xue, Nianwen",
    booktitle = "Proceedings of the 2024 Joint International Conference on Computational Linguistics, Language Resources and Evaluation (LREC-COLING 2024)",
    month = may,
    year = "2024",
    address = "Torino, Italia",
    publisher = "ELRA and ICCL",
    url = "https://aclanthology.org/2024.lrec-main.921/",
    pages = "10530--10538"
}

@inproceedings{li-etal-2017-dailydialog,
    title = "{D}aily{D}ialog: A Manually Labelled Multi-turn Dialogue Dataset",
    author = "Li, Yanran  and
      Su, Hui  and
      Shen, Xiaoyu  and
      Li, Wenjie  and
      Cao, Ziqiang  and
      Niu, Shuzi",
    editor = "Kondrak, Greg  and
      Watanabe, Taro",
    booktitle = "Proceedings of the Eighth International Joint Conference on Natural Language Processing (Volume 1: Long Papers)",
    month = nov,
    year = "2017",
    address = "Taipei, Taiwan",
    publisher = "Asian Federation of Natural Language Processing",
    url = "https://aclanthology.org/I17-1099/",
    pages = "986--995"
}

@inproceedings{zhang-etal-2018-personalizing-personachat,
    title = "Personalizing Dialogue Agents: {I} have a dog, do you have pets too?",
    author = "Zhang, Saizheng  and
      Dinan, Emily  and
      Urbanek, Jack  and
      Szlam, Arthur  and
      Kiela, Douwe  and
      Weston, Jason",
    editor = "Gurevych, Iryna  and
      Miyao, Yusuke",
    booktitle = "Proceedings of the 56th Annual Meeting of the Association for Computational Linguistics (Volume 1: Long Papers)",
    month = jul,
    year = "2018",
    address = "Melbourne, Australia",
    publisher = "Association for Computational Linguistics",
    url = "https://aclanthology.org/P18-1205/",
    doi = "10.18653/v1/P18-1205",
    pages = "2204--2213"
}

@inproceedings{budzianowski-etal-2018-multiwoz,
    title = "{M}ulti{WOZ} - A Large-Scale Multi-Domain {W}izard-of-{O}z Dataset for Task-Oriented Dialogue Modelling",
    author = "Budzianowski, Pawe{\l}  and
      Wen, Tsung-Hsien  and
      Tseng, Bo-Hsiang  and
      Casanueva, I{\~n}igo  and
      Ultes, Stefan  and
      Ramadan, Osman  and
      Ga{\v{s}}i{\'c}, Milica",
    editor = "Riloff, Ellen  and
      Chiang, David  and
      Hockenmaier, Julia  and
      Tsujii, Jun{'}ichi",
    booktitle = "Proceedings of the 2018 Conference on Empirical Methods in Natural Language Processing",
    month = oct # "-" # nov,
    year = "2018",
    address = "Brussels, Belgium",
    publisher = "Association for Computational Linguistics",
    url = "https://aclanthology.org/D18-1547/",
    doi = "10.18653/v1/D18-1547",
    pages = "5016--5026"
}

@inproceedings{liu-etal-2023-xdailydialog,
    title = "{XD}aily{D}ialog: A Multilingual Parallel Dialogue Corpus",
    author = "Liu, Zeming  and
      Nie, Ping  and
      Cai, Jie  and
      Wang, Haifeng  and
      Niu, Zheng-Yu  and
      Zhang, Peng  and
      Sachan, Mrinmaya  and
      Peng, Kaiping",
    editor = "Rogers, Anna  and
      Boyd-Graber, Jordan  and
      Okazaki, Naoaki",
    booktitle = "Proceedings of the 61st Annual Meeting of the Association for Computational Linguistics (Volume 1: Long Papers)",
    month = jul,
    year = "2023",
    address = "Toronto, Canada",
    publisher = "Association for Computational Linguistics",
    url = "https://aclanthology.org/2023.acl-long.684/",
    doi = "10.18653/v1/2023.acl-long.684",
    pages = "12240--12253"
}

@misc{zhang2022mdiabenchmarkmultilingualdialogue-mdia,
      title={MDIA: A Benchmark for Multilingual Dialogue Generation in 46 Languages}, 
      author={Qingyu Zhang and Xiaoyu Shen and Ernie Chang and Jidong Ge and Pengke Chen},
      year={2022},
      eprint={2208.13078},
      archivePrefix={arXiv},
      primaryClass={cs.CL},
      url={https://arxiv.org/abs/2208.13078}, 
}

@article{majewska-etal-2023-cross-cod,
    title = "Cross-Lingual Dialogue Dataset Creation via Outline-Based Generation",
    author = "Majewska, Olga  and
      Razumovskaia, Evgeniia  and
      Ponti, Edoardo M.  and
      Vuli{\'c}, Ivan  and
      Korhonen, Anna",
    journal = "Transactions of the Association for Computational Linguistics",
    volume = "11",
    year = "2023",
    address = "Cambridge, MA",
    publisher = "MIT Press",
    url = "https://aclanthology.org/2023.tacl-1.9/",
    doi = "10.1162/tacl_a_00539",
    pages = "139--156"
}

@inproceedings{kautsar-etal-2023-indotod,
    title = "{I}ndo{T}o{D}: A Multi-Domain {I}ndonesian Benchmark For End-to-End Task-Oriented Dialogue Systems",
    author = "Kautsar, Muhammad  and
      Nurdini, Rahmah  and
      Cahyawijaya, Samuel  and
      Winata, Genta  and
      Purwarianti, Ayu",
    editor = "Wijaya, Derry  and
      Aji, Alham Fikri  and
      Vania, Clara  and
      Winata, Genta Indra  and
      Purwarianti, Ayu",
    booktitle = "Proceedings of the First Workshop in South East Asian Language Processing",
    month = nov,
    year = "2023",
    address = "Nusa Dua, Bali, Indonesia",
    publisher = "Association for Computational Linguistics",
    url = "https://aclanthology.org/2023.sealp-1.7/",
    doi = "10.18653/v1/2023.sealp-1.7",
    pages = "85--99"
}

@misc{kautsar2025seadialoguesmultilingualculturallygrounded,
      title={SEADialogues: A Multilingual Culturally Grounded Multi-turn Dialogue Dataset on Southeast Asian Languages}, 
      author={Muhammad Dehan Al Kautsar and Aswin Candra and Muhammad Alif Al Hakim and Maxalmina Satria Kahfi and Fajri Koto and Alham Fikri Aji and Peerat Limkonchotiwat and Ekapol Chuangsuwanich and Genta Indra Winata},
      year={2025},
      eprint={2508.07069},
      archivePrefix={arXiv},
      primaryClass={cs.CL},
      url={https://arxiv.org/abs/2508.07069}, 
}

@inproceedings{aji-cohn-2025-loraxbench,
    title = "{LORAXBENCH}: A Multitask, Multilingual Benchmark Suite for 20 {I}ndonesian Languages",
    author = "Aji, Alham Fikri  and
      Cohn, Trevor",
    editor = "Christodoulopoulos, Christos  and
      Chakraborty, Tanmoy  and
      Rose, Carolyn  and
      Peng, Violet",
    booktitle = "Proceedings of the 2025 Conference on Empirical Methods in Natural Language Processing",
    month = nov,
    year = "2025",
    address = "Suzhou, China",
    publisher = "Association for Computational Linguistics",
    url = "https://aclanthology.org/2025.emnlp-main.881/",
    doi = "10.18653/v1/2025.emnlp-main.881",
    pages = "17421--17446",
    ISBN = "979-8-89176-332-6"
}

@inproceedings{levesque-2012-winograd,
author = {Levesque, Hector J. and Davis, Ernest and Morgenstern, Leora},
title = {The Winograd schema challenge},
year = {2012},
isbn = {9781577355601},
publisher = {AAAI Press},
booktitle = {Proceedings of the Thirteenth International Conference on Principles of Knowledge Representation and Reasoning},
pages = {552–561},
numpages = {10},
location = {Rome, Italy},
series = {KR'12},
url = {https://dl.acm.org/doi/10.5555/3031843.3031909}
}

@inproceedings{ponti-etal-2020-xcopa,
    title = "{XCOPA}: A Multilingual Dataset for Causal Commonsense Reasoning",
    author = "Ponti, Edoardo Maria  and
      Glava{\v{s}}, Goran  and
      Majewska, Olga  and
      Liu, Qianchu  and
      Vuli{\'c}, Ivan  and
      Korhonen, Anna",
    editor = "Webber, Bonnie  and
      Cohn, Trevor  and
      He, Yulan  and
      Liu, Yang",
    booktitle = "Proceedings of the 2020 Conference on Empirical Methods in Natural Language Processing (EMNLP)",
    month = nov,
    year = "2020",
    address = "Online",
    publisher = "Association for Computational Linguistics",
    url = "https://aclanthology.org/2020.emnlp-main.185/",
    doi = "10.18653/v1/2020.emnlp-main.185",
    pages = "2362--2376"
}

@inproceedings{lin-etal-2022-shot-xstorycloze,
    title = "Few-shot Learning with Multilingual Generative Language Models",
    author = "Lin, Xi Victoria  and
      Mihaylov, Todor  and
      Artetxe, Mikel  and
      Wang, Tianlu  and
      Chen, Shuohui  and
      Simig, Daniel  and
      Ott, Myle  and
      Goyal, Naman  and
      Bhosale, Shruti  and
      Du, Jingfei  and
      Pasunuru, Ramakanth  and
      Shleifer, Sam  and
      Koura, Punit Singh  and
      Chaudhary, Vishrav  and
      O{'}Horo, Brian  and
      Wang, Jeff  and
      Zettlemoyer, Luke  and
      Kozareva, Zornitsa  and
      Diab, Mona  and
      Stoyanov, Veselin  and
      Li, Xian",
    editor = "Goldberg, Yoav  and
      Kozareva, Zornitsa  and
      Zhang, Yue",
    booktitle = "Proceedings of the 2022 Conference on Empirical Methods in Natural Language Processing",
    month = dec,
    year = "2022",
    address = "Abu Dhabi, United Arab Emirates",
    publisher = "Association for Computational Linguistics",
    url = "https://aclanthology.org/2022.emnlp-main.616/",
    doi = "10.18653/v1/2022.emnlp-main.616",
    pages = "9019--9052"
}

@inproceedings{koto-etal-2022-indocloze,
    title = "Cloze Evaluation for Deeper Understanding of Commonsense Stories in {I}ndonesian",
    author = "Koto, Fajri  and
      Baldwin, Timothy  and
      Lau, Jey Han",
    editor = "Bosselut, Antoine  and
      Li, Xiang  and
      Lin, Bill Yuchen  and
      Shwartz, Vered  and
      Majumder, Bodhisattwa Prasad  and
      Lal, Yash Kumar  and
      Rudinger, Rachel  and
      Ren, Xiang  and
      Tandon, Niket  and
      Zouhar, Vil{\'e}m",
    booktitle = "Proceedings of the First Workshop on Commonsense Representation and Reasoning (CSRR 2022)",
    month = may,
    year = "2022",
    address = "Dublin, Ireland",
    publisher = "Association for Computational Linguistics",
    url = "https://aclanthology.org/2022.csrr-1.2/",
    doi = "10.18653/v1/2022.csrr-1.2",
    pages = "8--16"
}

@article{Muszyski2023AttentionCA,
  title={Attention checks and how to use them: Review and practical recommendations},
  author={Marek Muszyński},
  journal={Ask: Research and Methods},
  year={2023},
  url={https://api.semanticscholar.org/CorpusID:266098173},
  doi={10.18061/ask.v32i1.0001}
}

@inproceedings{kargaran-etal-2023-glotlid,
    title = "{G}lot{LID}: Language Identification for Low-Resource Languages",
    author = "Kargaran, Amir Hossein  and
      Imani, Ayyoob  and
      Yvon, Fran{\c{c}}ois  and
      Schuetze, Hinrich",
    editor = "Bouamor, Houda  and
      Pino, Juan  and
      Bali, Kalika",
    booktitle = "Findings of the Association for Computational Linguistics: EMNLP 2023",
    month = dec,
    year = "2023",
    address = "Singapore",
    publisher = "Association for Computational Linguistics",
    url = "https://aclanthology.org/2023.findings-emnlp.410/",
    doi = "10.18653/v1/2023.findings-emnlp.410",
    pages = "6155--6218"
}

@inproceedings{papineni-etal-2002-bleu,
    title = "{B}leu: a Method for Automatic Evaluation of Machine Translation",
    author = "Papineni, Kishore  and
      Roukos, Salim  and
      Ward, Todd  and
      Zhu, Wei-Jing",
    editor = "Isabelle, Pierre  and
      Charniak, Eugene  and
      Lin, Dekang",
    booktitle = "Proceedings of the 40th Annual Meeting of the Association for Computational Linguistics",
    month = jul,
    year = "2002",
    address = "Philadelphia, Pennsylvania, USA",
    publisher = "Association for Computational Linguistics",
    url = "https://aclanthology.org/P02-1040/",
    doi = "10.3115/1073083.1073135",
    pages = "311--318"
}

@inproceedings{zhang-etal-2020-bertscore,
  author       = {Tianyi Zhang and
                  Varsha Kishore and
                  Felix Wu and
                  Kilian Q. Weinberger and
                  Yoav Artzi},
  title        = {BERTScore: Evaluating Text Generation with {BERT}},
  booktitle    = {8th International Conference on Learning Representations, {ICLR} 2020,
                  Addis Ababa, Ethiopia, April 26-30, 2020},
  publisher    = {OpenReview.net},
  year         = {2020},
  url          = {https://openreview.net/forum?id=SkeHuCVFDr},
  bibsource    = {dblp computer science bibliography, https://dblp.org}
}

@inproceedings{novitasari-etal-2020-cross,
    title = "Cross-Lingual Machine Speech Chain for {J}avanese, {S}undanese, {B}alinese, and {B}ataks Speech Recognition and Synthesis",
    author = "Novitasari, Sashi  and
      Tjandra, Andros  and
      Sakti, Sakriani  and
      Nakamura, Satoshi",
    editor = "Beermann, Dorothee  and
      Besacier, Laurent  and
      Sakti, Sakriani  and
      Soria, Claudia",
    booktitle = "Proceedings of the 1st Joint Workshop on Spoken Language Technologies for Under-resourced languages (SLTU) and Collaboration and Computing for Under-Resourced Languages (CCURL)",
    month = may,
    year = "2020",
    address = "Marseille, France",
    publisher = "European Language Resources association",
    url = "https://aclanthology.org/2020.sltu-1.18/",
    pages = "131--138",
    language = "eng",
    ISBN = "979-10-95546-35-1"
}

@misc{permadi-et-al-2026-idmocqa,
      title={No Shortcuts to Culture: Indonesian Multi-hop Question Answering for Complex Cultural Understanding}, 
      author={Vynska Amalia Permadi and Xingwei Tan and Nafise Sadat Moosavi and Nikos Aletras},
      year={2026},
      eprint={2602.03709},
      archivePrefix={arXiv},
      primaryClass={cs.CL},
      url={https://arxiv.org/abs/2602.03709}, 
}

@misc{grattafiori2024llama3herdmodels,
      title={The {L}lama 3 Herd of Models}, 
      author={Aaron Grattafiori and Abhimanyu Dubey and Abhinav Jauhri and Abhinav Pandey and Abhishek Kadian and Ahmad Al-Dahle and Aiesha Letman and Akhil Mathur and Alan Schelten and Alex Vaughan and Amy Yang and Angela Fan and Anirudh Goyal and Anthony Hartshorn and Aobo Yang and Archi Mitra and Archie Sravankumar and Artem Korenev and Arthur Hinsvark and Arun Rao and Aston Zhang and Aurelien Rodriguez and Austen Gregerson and Ava Spataru and Baptiste Roziere and Bethany Biron and Binh Tang and Bobbie Chern and Charlotte Caucheteux and Chaya Nayak and Chloe Bi and Chris Marra and Chris McConnell and Christian Keller and Christophe Touret and Chunyang Wu and Corinne Wong and Cristian Canton Ferrer and Cyrus Nikolaidis and Damien Allonsius and Daniel Song and Danielle Pintz and Danny Livshits and Danny Wyatt and David Esiobu and Dhruv Choudhary and Dhruv Mahajan and Diego Garcia-Olano and Diego Perino and Dieuwke Hupkes and Egor Lakomkin and Ehab AlBadawy and Elina Lobanova and Emily Dinan and Eric Michael Smith and Filip Radenovic and Francisco Guzmán and Frank Zhang and Gabriel Synnaeve and Gabrielle Lee and Georgia Lewis Anderson and Govind Thattai and Graeme Nail and Gregoire Mialon and Guan Pang and Guillem Cucurell and Hailey Nguyen and Hannah Korevaar and Hu Xu and Hugo Touvron and Iliyan Zarov and Imanol Arrieta Ibarra and Isabel Kloumann and Ishan Misra and Ivan Evtimov and Jack Zhang and Jade Copet and Jaewon Lee and Jan Geffert and Jana Vranes and Jason Park and Jay Mahadeokar and Jeet Shah and Jelmer van der Linde and Jennifer Billock and Jenny Hong and Jenya Lee and Jeremy Fu and Jianfeng Chi and Jianyu Huang and Jiawen Liu and Jie Wang and Jiecao Yu and Joanna Bitton and Joe Spisak and Jongsoo Park and Joseph Rocca and Joshua Johnstun and Joshua Saxe and Junteng Jia and Kalyan Vasuden Alwala and Karthik Prasad and Kartikeya Upasani and Kate Plawiak and Ke Li and Kenneth Heafield and Kevin Stone and Khalid El-Arini and Krithika Iyer and Kshitiz Malik and Kuenley Chiu and Kunal Bhalla and Kushal Lakhotia and Lauren Rantala-Yeary and Laurens van der Maaten and Lawrence Chen and Liang Tan and Liz Jenkins and Louis Martin and Lovish Madaan and Lubo Malo and Lukas Blecher and Lukas Landzaat and Luke de Oliveira and Madeline Muzzi and Mahesh Pasupuleti and Mannat Singh and Manohar Paluri and Marcin Kardas and Maria Tsimpoukelli and Mathew Oldham and Mathieu Rita and Maya Pavlova and Melanie Kambadur and Mike Lewis and Min Si and Mitesh Kumar Singh and Mona Hassan and Naman Goyal and Narjes Torabi and Nikolay Bashlykov and Nikolay Bogoychev and Niladri Chatterji and Ning Zhang and Olivier Duchenne and Onur Çelebi and Patrick Alrassy and Pengchuan Zhang and Pengwei Li and Petar Vasic and Peter Weng and Prajjwal Bhargava and Pratik Dubal and Praveen Krishnan and Punit Singh Koura and Puxin Xu and Qing He and Qingxiao Dong and Ragavan Srinivasan and Raj Ganapathy and Ramon Calderer and Ricardo Silveira Cabral and Robert Stojnic and Roberta Raileanu and Rohan Maheswari and Rohit Girdhar and Rohit Patel and Romain Sauvestre and Ronnie Polidoro and Roshan Sumbaly and Ross Taylor and Ruan Silva and Rui Hou and Rui Wang and Saghar Hosseini and Sahana Chennabasappa and Sanjay Singh and Sean Bell and Seohyun Sonia Kim and Sergey Edunov and Shaoliang Nie and Sharan Narang and Sharath Raparthy and Sheng Shen and Shengye Wan and Shruti Bhosale and Shun Zhang and Simon Vandenhende and Soumya Batra and Spencer Whitman and Sten Sootla and Stephane Collot and Suchin Gururangan and Sydney Borodinsky and Tamar Herman and Tara Fowler and Tarek Sheasha and Thomas Georgiou and Thomas Scialom and Tobias Speckbacher and Todor Mihaylov and Tong Xiao and Ujjwal Karn and Vedanuj Goswami and Vibhor Gupta and Vignesh Ramanathan and Viktor Kerkez and Vincent Gonguet and Virginie Do and Vish Vogeti and Vítor Albiero and Vladan Petrovic and Weiwei Chu and Wenhan Xiong and Wenyin Fu and Whitney Meers and Xavier Martinet and Xiaodong Wang and Xiaofang Wang and Xiaoqing Ellen Tan and Xide Xia and Xinfeng Xie and Xuchao Jia and Xuewei Wang and Yaelle Goldschlag and Yashesh Gaur and Yasmine Babaei and Yi Wen and Yiwen Song and Yuchen Zhang and Yue Li and Yuning Mao and Zacharie Delpierre Coudert and Zheng Yan and Zhengxing Chen and Zoe Papakipos and Aaditya Singh and Aayushi Srivastava and Abha Jain and Adam Kelsey and Adam Shajnfeld and Adithya Gangidi and Adolfo Victoria and Ahuva Goldstand and Ajay Menon and Ajay Sharma and Alex Boesenberg and Alexei Baevski and Allie Feinstein and Amanda Kallet and Amit Sangani and Amos Teo and Anam Yunus and Andrei Lupu and Andres Alvarado and Andrew Caples and Andrew Gu and Andrew Ho and Andrew Poulton and Andrew Ryan and Ankit Ramchandani and Annie Dong and Annie Franco and Anuj Goyal and Aparajita Saraf and Arkabandhu Chowdhury and Ashley Gabriel and Ashwin Bharambe and Assaf Eisenman and Azadeh Yazdan and Beau James and Ben Maurer and Benjamin Leonhardi and Bernie Huang and Beth Loyd and Beto De Paola and Bhargavi Paranjape and Bing Liu and Bo Wu and Boyu Ni and Braden Hancock and Bram Wasti and Brandon Spence and Brani Stojkovic and Brian Gamido and Britt Montalvo and Carl Parker and Carly Burton and Catalina Mejia and Ce Liu and Changhan Wang and Changkyu Kim and Chao Zhou and Chester Hu and Ching-Hsiang Chu and Chris Cai and Chris Tindal and Christoph Feichtenhofer and Cynthia Gao and Damon Civin and Dana Beaty and Daniel Kreymer and Daniel Li and David Adkins and David Xu and Davide Testuggine and Delia David and Devi Parikh and Diana Liskovich and Didem Foss and Dingkang Wang and Duc Le and Dustin Holland and Edward Dowling and Eissa Jamil and Elaine Montgomery and Eleonora Presani and Emily Hahn and Emily Wood and Eric-Tuan Le and Erik Brinkman and Esteban Arcaute and Evan Dunbar and Evan Smothers and Fei Sun and Felix Kreuk and Feng Tian and Filippos Kokkinos and Firat Ozgenel and Francesco Caggioni and Frank Kanayet and Frank Seide and Gabriela Medina Florez and Gabriella Schwarz and Gada Badeer and Georgia Swee and Gil Halpern and Grant Herman and Grigory Sizov and Guangyi and Zhang and Guna Lakshminarayanan and Hakan Inan and Hamid Shojanazeri and Han Zou and Hannah Wang and Hanwen Zha and Haroun Habeeb and Harrison Rudolph and Helen Suk and Henry Aspegren and Hunter Goldman and Hongyuan Zhan and Ibrahim Damlaj and Igor Molybog and Igor Tufanov and Ilias Leontiadis and Irina-Elena Veliche and Itai Gat and Jake Weissman and James Geboski and James Kohli and Janice Lam and Japhet Asher and Jean-Baptiste Gaya and Jeff Marcus and Jeff Tang and Jennifer Chan and Jenny Zhen and Jeremy Reizenstein and Jeremy Teboul and Jessica Zhong and Jian Jin and Jingyi Yang and Joe Cummings and Jon Carvill and Jon Shepard and Jonathan McPhie and Jonathan Torres and Josh Ginsburg and Junjie Wang and Kai Wu and Kam Hou U and Karan Saxena and Kartikay Khandelwal and Katayoun Zand and Kathy Matosich and Kaushik Veeraraghavan and Kelly Michelena and Keqian Li and Kiran Jagadeesh and Kun Huang and Kunal Chawla and Kyle Huang and Lailin Chen and Lakshya Garg and Lavender A and Leandro Silva and Lee Bell and Lei Zhang and Liangpeng Guo and Licheng Yu and Liron Moshkovich and Luca Wehrstedt and Madian Khabsa and Manav Avalani and Manish Bhatt and Martynas Mankus and Matan Hasson and Matthew Lennie and Matthias Reso and Maxim Groshev and Maxim Naumov and Maya Lathi and Meghan Keneally and Miao Liu and Michael L. Seltzer and Michal Valko and Michelle Restrepo and Mihir Patel and Mik Vyatskov and Mikayel Samvelyan and Mike Clark and Mike Macey and Mike Wang and Miquel Jubert Hermoso and Mo Metanat and Mohammad Rastegari and Munish Bansal and Nandhini Santhanam and Natascha Parks and Natasha White and Navyata Bawa and Nayan Singhal and Nick Egebo and Nicolas Usunier and Nikhil Mehta and Nikolay Pavlovich Laptev and Ning Dong and Norman Cheng and Oleg Chernoguz and Olivia Hart and Omkar Salpekar and Ozlem Kalinli and Parkin Kent and Parth Parekh and Paul Saab and Pavan Balaji and Pedro Rittner and Philip Bontrager and Pierre Roux and Piotr Dollar and Polina Zvyagina and Prashant Ratanchandani and Pritish Yuvraj and Qian Liang and Rachad Alao and Rachel Rodriguez and Rafi Ayub and Raghotham Murthy and Raghu Nayani and Rahul Mitra and Rangaprabhu Parthasarathy and Raymond Li and Rebekkah Hogan and Robin Battey and Rocky Wang and Russ Howes and Ruty Rinott and Sachin Mehta and Sachin Siby and Sai Jayesh Bondu and Samyak Datta and Sara Chugh and Sara Hunt and Sargun Dhillon and Sasha Sidorov and Satadru Pan and Saurabh Mahajan and Saurabh Verma and Seiji Yamamoto and Sharadh Ramaswamy and Shaun Lindsay and Shaun Lindsay and Sheng Feng and Shenghao Lin and Shengxin Cindy Zha and Shishir Patil and Shiva Shankar and Shuqiang Zhang and Shuqiang Zhang and Sinong Wang and Sneha Agarwal and Soji Sajuyigbe and Soumith Chintala and Stephanie Max and Stephen Chen and Steve Kehoe and Steve Satterfield and Sudarshan Govindaprasad and Sumit Gupta and Summer Deng and Sungmin Cho and Sunny Virk and Suraj Subramanian and Sy Choudhury and Sydney Goldman and Tal Remez and Tamar Glaser and Tamara Best and Thilo Koehler and Thomas Robinson and Tianhe Li and Tianjun Zhang and Tim Matthews and Timothy Chou and Tzook Shaked and Varun Vontimitta and Victoria Ajayi and Victoria Montanez and Vijai Mohan and Vinay Satish Kumar and Vishal Mangla and Vlad Ionescu and Vlad Poenaru and Vlad Tiberiu Mihailescu and Vladimir Ivanov and Wei Li and Wenchen Wang and Wenwen Jiang and Wes Bouaziz and Will Constable and Xiaocheng Tang and Xiaojian Wu and Xiaolan Wang and Xilun Wu and Xinbo Gao and Yaniv Kleinman and Yanjun Chen and Ye Hu and Ye Jia and Ye Qi and Yenda Li and Yilin Zhang and Ying Zhang and Yossi Adi and Youngjin Nam and Yu and Wang and Yu Zhao and Yuchen Hao and Yundi Qian and Yunlu Li and Yuzi He and Zach Rait and Zachary DeVito and Zef Rosnbrick and Zhaoduo Wen and Zhenyu Yang and Zhiwei Zhao and Zhiyu Ma},
      year={2024},
      eprint={2407.21783},
      archivePrefix={arXiv},
      primaryClass={cs.AI},
      url={https://arxiv.org/abs/2407.21783}, 
}

@article{qwen3technicalreport,
  title={Qwen3 technical report},
  author={Yang, An and Li, Anfeng and Yang, Baosong and Zhang, Beichen and Hui, Binyuan and Zheng, Bo and Yu, Bowen and Gao, Chang and Huang, Chengen and Lv, Chenxu and others},
  journal={arXiv preprint arXiv:2505.09388},
  year={2025},
  url={https://arxiv.org/abs/2505.09388}
}

@misc{gemma_2024,
      title={{G}emma: Open Models Based on Gemini Research and Technology}, 
      author={Gemma Team and Thomas Mesnard and Cassidy Hardin and Robert Dadashi and Surya Bhupatiraju and Shreya Pathak and Laurent Sifre and Morgane Rivière and Mihir Sanjay Kale and Juliette Love and Pouya Tafti and Léonard Hussenot and Pier Giuseppe Sessa and Aakanksha Chowdhery and Adam Roberts and Aditya Barua and Alex Botev and Alex Castro-Ros and Ambrose Slone and Amélie Héliou and Andrea Tacchetti and Anna Bulanova and Antonia Paterson and Beth Tsai and Bobak Shahriari and Charline Le Lan and Christopher A. Choquette-Choo and Clément Crepy and Daniel Cer and Daphne Ippolito and David Reid and Elena Buchatskaya and Eric Ni and Eric Noland and Geng Yan and George Tucker and George-Christian Muraru and Grigory Rozhdestvenskiy and Henryk Michalewski and Ian Tenney and Ivan Grishchenko and Jacob Austin and James Keeling and Jane Labanowski and Jean-Baptiste Lespiau and Jeff Stanway and Jenny Brennan and Jeremy Chen and Johan Ferret and Justin Chiu and Justin Mao-Jones and Katherine Lee and Kathy Yu and Katie Millican and Lars Lowe Sjoesund and Lisa Lee and Lucas Dixon and Machel Reid and Maciej Mikuła and Mateo Wirth and Michael Sharman and Nikolai Chinaev and Nithum Thain and Olivier Bachem and Oscar Chang and Oscar Wahltinez and Paige Bailey and Paul Michel and Petko Yotov and Rahma Chaabouni and Ramona Comanescu and Reena Jana and Rohan Anil and Ross McIlroy and Ruibo Liu and Ryan Mullins and Samuel L Smith and Sebastian Borgeaud and Sertan Girgin and Sholto Douglas and Shree Pandya and Siamak Shakeri and Soham De and Ted Klimenko and Tom Hennigan and Vlad Feinberg and Wojciech Stokowiec and Yu-hui Chen and Zafarali Ahmed and Zhitao Gong and Tris Warkentin and Ludovic Peran and Minh Giang and Clément Farabet and Oriol Vinyals and Jeff Dean and Koray Kavukcuoglu and Demis Hassabis and Zoubin Ghahramani and Douglas Eck and Joelle Barral and Fernando Pereira and Eli Collins and Armand Joulin and Noah Fiedel and Evan Senter and Alek Andreev and Kathleen Kenealy},
      year={2024},
      eprint={2403.08295},
      archivePrefix={arXiv},
      primaryClass={cs.CL},
      url={https://arxiv.org/abs/2403.08295}, 
}

@misc{dou2025sailor2sailingsoutheastasia,
      title={Sailor2: Sailing in South-East Asia with Inclusive Multilingual LLMs}, 
      author={Longxu Dou and Qian Liu and Fan Zhou and Changyu Chen and Zili Wang and Ziqi Jin and Zichen Liu and Tongyao Zhu and Cunxiao Du and Penghui Yang and Haonan Wang and Jiaheng Liu and Yongchi Zhao and Xiachong Feng and Xin Mao and Man Tsung Yeung and Kunat Pipatanakul and Fajri Koto and Min Si Thu and Hynek Kydlíček and Zeyi Liu and Qunshu Lin and Sittipong Sripaisarnmongkol and Kridtaphad Sae-Khow and Nirattisai Thongchim and Taechawat Konkaew and Narong Borijindargoon and Anh Dao and Matichon Maneegard and Phakphum Artkaew and Zheng-Xin Yong and Quan Nguyen and Wannaphong Phatthiyaphaibun and Hoang H. Tran and Mike Zhang and Shiqi Chen and Tianyu Pang and Chao Du and Xinyi Wan and Wei Lu and Min Lin},
      year={2025},
      eprint={2502.12982},
      archivePrefix={arXiv},
      primaryClass={cs.CL},
      url={https://arxiv.org/abs/2502.12982}, 
}

@misc{SEA-LION-2504.05747,
      title={SEA-LION: Southeast Asian Languages in One Network},
      author={Raymond Ng and Thanh Ngan Nguyen and Yuli Huang and Ngee Chia Tai and Wai Yi Leong and Wei Qi Leong and Xianbin Yong and Jian Gang Ngui and Yosephine Susanto and Nicholas Cheng and Hamsawardhini Rengarajan and Peerat Limkonchotiwat and Adithya Venkatadri Hulagadri and Kok Wai Teng and Yeo Yeow Tong and Bryan Siow and Wei Yi Teo and Wayne Lau and Choon Meng Tan and Brandon Ong and Zhi Hao Ong and Jann Railey Montalan and Adwin Chan and Sajeban Antonyrex and Ren Lee and Esther Choa and David Ong Tat-Wee and Bing Jie Darius Liu and William Chandra Tjhi and Erik Cambria and Leslie Teo},
      year={2025},
      eprint={2504.05747},
      archivePrefix={arXiv},
      primaryClass={cs.CL},
      url={https://arxiv.org/abs/2504.05747},
}

\appendix

\newpage

\section{Dataset Creation Additional Details}
\label{sec:appendix-dialogue-generation}

\begin{figure*}[t]
  \centering
  \includegraphics[width=1.9\columnwidth]{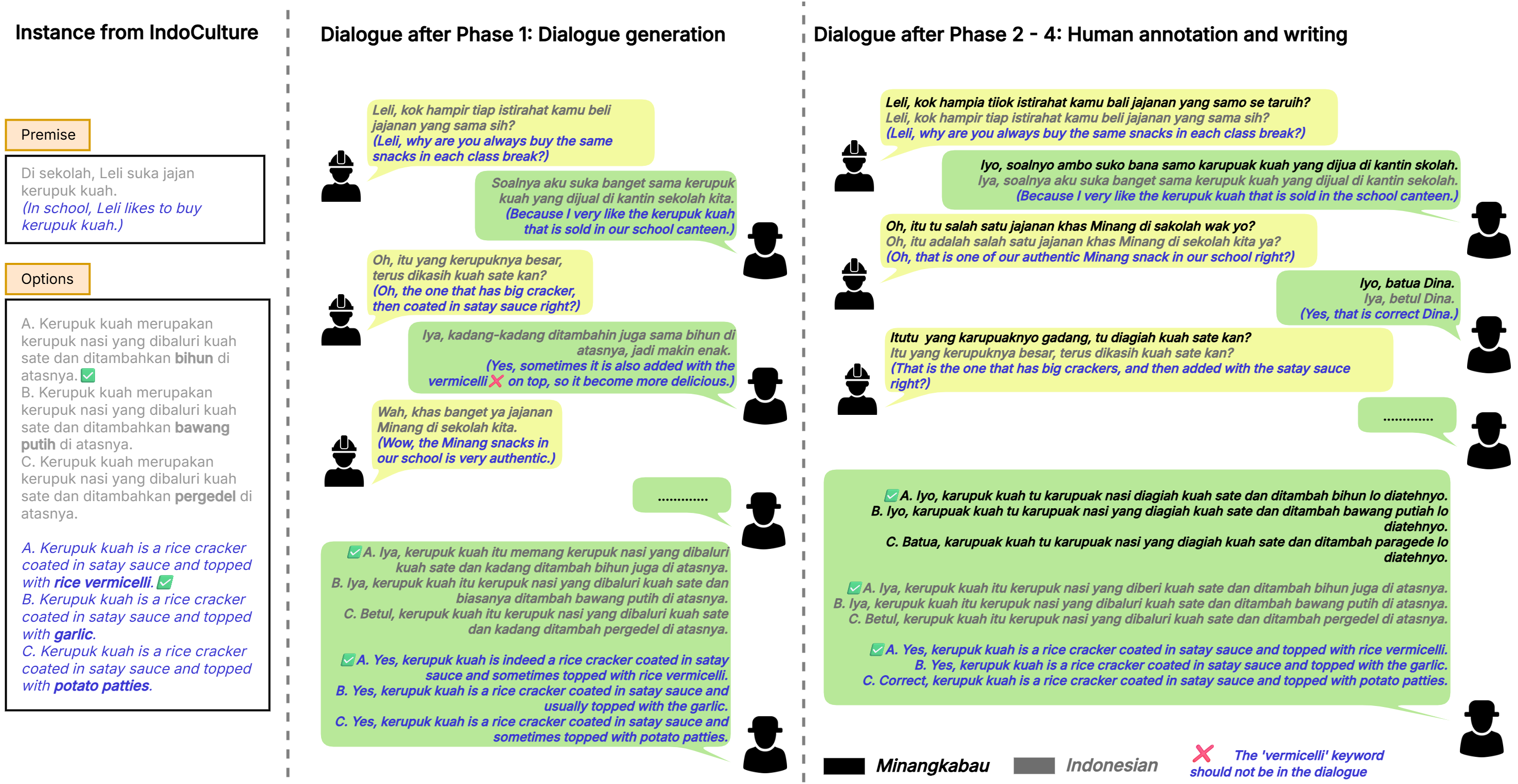}
  \caption{An example of the dataset transformation process from an original IndoCulture instance into \indoculturedialoguetitle\ across each phase. This example uses the Minangkabau cultural setting.}
  \label{fig:appendix-dialogue-generation-transformation}
\end{figure*}

As mentioned, we repurpose the IndoCulture~\cite{koto-etal-2024-indoculture} and COPAL-ID~\cite{wibowo-etal-2024-copal} as both are grounded in Indonesian cultural contexts and share a similar structure: multiple-choice questions based on culturally situated scenarios, where each instance consists of one premise and three answer options with a single correct answer. This structure makes them well-suited for transformation into a unified dialogue dataset. Notably, IndoCulture contains both province-specific and non-province-specific instances, whereas COPAL-ID focuses specifically on Jakarta-related cultural contexts. Using GPT-5, we generate dialogues by treating the premise as the dialogue context and the three answer options as candidate final utterances, which are then passed to annotators for validation and refinement. The prompt used for dialogue generation is shown in Figure~\ref{fig:appendix-prompt-dialogue-generation}, while the transformation process is illustrated in Figure~\ref{fig:appendix-dialogue-generation-transformation}.

\begin{figure}[!h]
    \centering
\todo[inline, color=gray!20]
{ \small 
You are a dialogue creation agent, and you have the task of creating a daily-life dialogue in Indonesian between two people, with a maximum length of 6 sentences. The dialogue must contain cultural elements from the region of \{province\}, and relate to the topic of \{topic\}. Within the dialogue, there must be an implied \{context\}, and it should end with one 'correct utterance' that matches the correct \{answer\} and two 'incorrect utterances' from all of the options in \{option\} except the {answer}. Do not mention the answer nor state it explicitly within the dialogue, as it must only occur on the one 'correct utterance'! It is also important that the relationship between the context and the answer reflects commonsense knowledge that is typical of the region.

\vspace{6pt}
Please create a dialogue based on these variables using markdown format, so it can be extracted easily in the postprocess.

\vspace{6pt}
==FEW SHOT EXAMPLES==

\vspace{6pt}
\{3\_fewshot\_examples\}

\vspace{6pt}
==DIALOGUE==

\vspace{6pt}
\{dialogue\}

}
    \caption{Prompt for the dialogue generation.}
    \label{fig:appendix-prompt-dialogue-generation}
\end{figure}

\section{Detailed Dataset Statistics}
\label{sec:appendix-detailed-data-stats}

Table~\ref{tab:appendix-data-stats-region} and Table~\ref{tab:appendix-data-stats-topic} present detailed dataset statistics by region and topic, respectively. The dataset covers 10 provinces and one general Indonesian cultural category, with the largest number of dialogues originating from the Minangkabau culture in West Sumatra and the fewest from NTT. Across topics, food-related dialogues are the most common, while agriculture-related dialogues are the least represented. We also observe that local-language dialogues generally contain higher word counts and lexical diversity across both regions and topics.

\setlength{\tabcolsep}{0.12em}
\begin{table*}[h]
\centering
\small
\resizebox{\textwidth}{!}{%
\begin{tabular}{l | ccccccccccc}
\toprule
\textbf{Metric} 
& \textbf{Indonesia} & \textbf{Aceh} & \textbf{Bali} & \textbf{W. Java} & \textbf{C. Java} & \textbf{E. Java} & \textbf{S. Kalimantan} & \textbf{NTT} & \textbf{W. Papua} & \textbf{S. Sulawesi} & \textbf{W. Sumatera} \\
\midrule

\multicolumn{12}{c}{\textbf{General Dialogue Statistics}} \\
\#dialogues 
& 500 & 492 & 482 & 462 & 466 & 342 & 466 & 34 & 506 & 148 & 598 \\
\#region specific dialogues 
& 0 & 348 & 368 & 268 & 324 & 228 & 390 & 28 & 446 & 134 & 446 \\
\midrule

\multicolumn{12}{c}{\textbf{Indonesian Data}} \\
avg. words per dialogue 
& 52.27 & 59.80 & 60.21 & 60.04 & 59.86 & 59.68 & 61.19 & 60.06 & 62.04 & 60.26 & 59.08 \\
avg. utterances per dialogue 
& 6.02 & 6.96 & 6.22 & 7.73 & 6.61 & 6.81 & 6.40 & 6.88 & 6.45 & 6.23 & 6.63 \\
avg. words per utterance 
& 7.72 & 7.73 & 8.72 & 7.00 & 8.15 & 7.90 & 8.63 & 7.85 & 8.70 & 8.71 & 8.03 \\
\#word counts 
& 13,068 & 14,710 & 14,511 & 13,870 & 13,947 & 10,205 & 14,257 & 1,021 & 15,695 & 4,459 & 17,666 \\
\#unique word counts 
& 2,895 & 2,936 & 2,974 & 3,157 & 2,867 & 2,563 & 2,958 & 481 & 3,063 & 1,383 & 3,636 \\
\midrule

\multicolumn{12}{c}{\textbf{Local Language Data}} \\
avg. words per dialogue 
& 52.27 & 63.52 & 65.12 & 61.21 & 60.85 & 60.69 & 62.68 & 67.94 & 62.53 & 65.85 & 59.70 \\
avg. utterances per dialogue 
& 6.02 & 6.96 & 6.22 & 7.73 & 6.61 & 6.81 & 6.40 & 6.88 & 6.45 & 6.23 & 6.63 \\
avg. words per utterance 
& 7.72 & 8.27 & 10.06 & 7.15 & 8.38 & 8.17 & 8.86 & 9.76 & 8.79 & 9.68 & 8.75 \\
\#word counts 
& 13,068 & 15,626 & 15,695 & 14,140 & 14,177 & 10,378 & 14,604 & 1,155 & 15,819 & 4,873 & 17,849 \\
\#unique word counts 
& 2,895 & 3,388 & 3,905 & 3,400 & 3,469 & 3,052 & 3,329 & 499 & 3,156 & 1,664 & 3,868 \\
\bottomrule
\end{tabular}%
}
\caption{Detailed dataset statistics, split by region. W, C, E, and S denote west, central, east, and south, respectively.}
\label{tab:appendix-data-stats-region}
\end{table*}

\setlength{\tabcolsep}{0.12em}
\begin{table*}[t]
\centering
\small
\resizebox{\textwidth}{!}{%
\begin{tabular}{l | cccccccccccc}
\toprule
\textbf{Metric} 
& \textbf{Food} & \textbf{Wedding} & \textbf{Family} & \textbf{Birth} & \textbf{Death} & \textbf{Religion} & \textbf{Agriculture} & \textbf{Farm} & \textbf{Art} & \textbf{Game} & \textbf{Daily} & \textbf{Socio-religious} \\
\midrule

\multicolumn{13}{c}{\textbf{General Dialogue Statistics (By Topic)}} \\
\#dialogues 
& 646 & 522 & 344 & 424 & 258 & 300 & 154 & 170 & 424 & 158 & 246 & 350 \\
\#country specific dialogues 
& 544 & 426 & 246 & 264 & 162 & 186 & 106 & 126 & 398 & 134 & 146 & 242 \\
\midrule

\multicolumn{13}{c}{\textbf{Indonesian Data}} \\
avg. words per dialogue 
& 58.31 & 61.07 & 59.87 & 61.31 & 61.19 & 61.99 & 59.48 & 60.92 & 61.07 & 58.14 & 57.93 & 60.91 \\
avg. utterances per dialogue 
& 6.47 & 6.66 & 6.66 & 6.98 & 7.13 & 6.67 & 7.47 & 7.45 & 6.29 & 6.03 & 6.78 & 6.63 \\
avg. words per utterance 
& 8.09 & 8.29 & 8.10 & 7.93 & 7.75 & 8.40 & 7.16 & 7.38 & 8.76 & 8.67 & 7.67 & 8.29 \\
\#word counts 
& 18,834 & 15,938 & 10,298 & 12,997 & 7,893 & 9,298 & 4,580 & 5,178 & 12,947 & 4,593 & 7,125 & 10,660 \\
\#unique word counts 
& 3,183 & 2,875 & 2,434 & 2,592 & 1,725 & 2,021 & 1,308 & 1,570 & 2,690 & 1,316 & 1,992 & 2,732 \\
\midrule

\multicolumn{13}{c}{\textbf{Local Language Data}} \\
avg. words per dialogue 
& 61.40 & 63.78 & 62.16 & 63.10 & 62.77 & 63.41 & 61.43 & 62.34 & 62.30 & 59.18 & 59.41 & 62.54 \\
avg. utterances per dialogue 
& 6.35 & 6.50 & 6.53 & 6.97 & 7.12 & 6.57 & 7.38 & 7.32 & 6.21 & 5.92 & 6.69 & 6.60 \\
avg. words per utterance 
& 8.89 & 8.99 & 8.68 & 8.25 & 8.02 & 8.80 & 7.56 & 7.74 & 9.10 & 9.04 & 8.04 & 8.62 \\
\#word counts 
& 19,831 & 16,646 & 10,691 & 13,378 & 8,097 & 9,511 & 4,730 & 5,299 & 13,207 & 4,675 & 7,307 & 10,944 \\
\#unique word counts 
& 6,361 & 5,385 & 4,181 & 4,697 & 3,092 & 3,457 & 2,151 & 2,486 & 4,381 & 2,142 & 3,288 & 4,562 \\
\bottomrule
\end{tabular}%
}
\caption{Detailed dataset statistics by topic.}
\label{tab:appendix-data-stats-topic}
\end{table*}

\section{\indoculturedialoguetitle\ Tasks Prompt}
\label{sec:appendix-task-prompt}

We present the prompts used for evaluation in Figure~\ref{fig:appendix-prompt-mcq},~\ref{fig:appendix-system-prompt-mt},~\ref{fig:appendix-user-prompt-mt-indo-local},~\ref{fig:appendix-user-prompt-mt-local-to-indo}, and~\ref{fig:appendix-prompt-dialogue-steer}. To ensure fairness, we use the same system and user prompts across all models for each task. For the MCQ task, province and language information can be optionally included to control the amount of contextual information provided, resulting in the different settings reported in Table~\ref{tab:main-local-dialogue-delta}. For open-source models, we adopt a likelihood-based evaluation setup by computing the likelihood of each answer option (A, B, or C) and selecting the highest-probability choice. This approach provides a more stable evaluation than generation-based prompting, which can be highly sensitive to prompt variations, especially for smaller models.

For the machine translation and language steering tasks, the same prompt templates are used across all model categories, with only the language and province variables adjusted according to the target setting.

\begin{figure}[!h]
  \centering
\todo[inline, color=yellow!20]
{\small You are an Indonesian professional that able to reason in the Indonesian culture.

\vspace{6pt}
Rules:

1. Output only the OPTIONS [A/B/C].

2. Do not add explanations, comments, or quotation marks. Only the option label [A/B/C].

\vspace{6pt}
You are tasked with selecting the most culturally appropriate option based on the context provided below.

\vspace{6pt}
Location: \{province\}

Language: \{language\}

\vspace{6pt}
Conversation:

\{dialogue\}

\vspace{6pt}
Give the option label only [A/B/C]. 

\vspace{6pt}
Options:

\{choices\_str\}

\vspace{6pt}
Answer:
}
  \caption{Task 1 - Dialogue-based cultural commonsense MCQ Evaluation's prompt.}
  \label{fig:appendix-prompt-mcq}
\end{figure}

\begin{figure}[!h]
    \centering
\todo[inline, color=purple!20]
{ \small You are a professional Indonesian translation system, specialized in 
    Indonesian and Indonesian local languages.

    \vspace{6pt}
    Rules:
    
    1. Output only the translated text in Latin (both Indonesian and other local languages).
    
    2. Do not add explanations, comments, labels, or quotation marks.
    
    3. Preserve meaning, tone, and level of formality.
    
    4. Do not change names, numbers, or entities.
    
    5. If the input already matches the target variety, return it unchanged.
    
    \vspace{6pt}
    Extra Rules:
    
    - Output ONLY the translation
    
    - No explanation
    
    - No reasoning (DO NOT EXPLAIN YOUR REASONING)
    
    - No extra text
}    
    \caption{Task 2 - Machine translation's unified system prompt.}
    \label{fig:appendix-system-prompt-mt}
\end{figure}

\begin{figure}[!h]
    \centering
\todo[inline, color=purple!20]
{ \small Translate the following Indonesian text into the \{language\}, the local language of \{province\}. Do not include any explanation, just return the \{language\} translation.

    \vspace{6pt}
    Indonesian Text: \{indonesian\_dialogue\}
    
    \{language\} Translation:
}    
    \caption{Task 2 - Machine translation's (Indonesian $\rightarrow$ local Indonesian) user prompt.}
    \label{fig:appendix-user-prompt-mt-indo-local}
\end{figure}

\begin{figure}[!h]
    \centering
\todo[inline, color=purple!20]
{ \small Translate the following \{language\} text from the \{province\} into Indonesian. Do not include any explanation, just return the Indonesian translation.

    \vspace{6pt}
    \{language\} Text: \{local\_dialogue\}
    
    Indonesian Translation:
}    
    \caption{Task 2 - Machine translation's (local Indonesian $\rightarrow$  Indonesian) user prompt.}
    \label{fig:appendix-user-prompt-mt-local-to-indo}
\end{figure}

\begin{figure}[!h]
    \centering
\todo[inline, color=blue!20]
{ \small You are a helpful assistant who writes in \{target\_name\}\{code\_suffix\}. Continue the dialogue with a single natural utterance in that variety.

\vspace{6pt}
Rules: output ONLY the continuation text (no analysis, no translation, no English explanation, no meta commentary, no speaker labels).

\vspace{6pt}
Do NOT output <think> blocks or any reasoning traces.
}    
    \caption{Task 3 - Language steering's user prompt.}
    \label{fig:appendix-prompt-dialogue-steer}
\end{figure}

\section{Language Covered by \indoculturedialoguetitle}
\label{sec:appendix-language-covered-indoculturedialogue}

We provide details of the covered provinces, languages, and estimated speaker populations in Table~\ref{tab:appendix-language-covered}. Our benchmark covers 10 provinces and one general Indonesian cultural category. Javanese is further divided into three dialects: Dermayu, primarily spoken in eastern West Java; Mataraman, spoken in Central Java; and Arekan, spoken in Surabaya (East Java) and surrounding regions.

The estimated speaker populations highlight the importance of evaluating cultural understanding in LLMs for these languages, many of which have millions of speakers. Despite this, language technology support for several local languages remains limited, and some languages, such as Acehnese, Minangkabau, and Wamesa, are considered endangered languages.

\setlength{\tabcolsep}{0.12em}
\begin{table}[!h]
\centering
\small
\resizebox{0.47\textwidth}{!}{%
\begin{tabular}{llc}
\toprule
\textbf{Province} & \textbf{Language Covered (code)} & \textbf{Estimated Speaker} \\
\midrule
West Java & Javanese - Dermayu dialect (\textit{jav}) & \\
Central Java & Javanese - Mataraman dialect (\textit{jav}) & 91,000,000 * \\
East Java & Javanese - Arekan dialect (\textit{jav}) &  \\
\hline
Aceh & Acehnese (\textit{ace}) & 3,700,000 *** \\
West Sumatra & Minangkabau (\textit{min}) & 8,000,000 *** \\
Bali & Balinese (\textit{ban}) & 4,800,000 ** \\
South Kalimantan & Banjarese (\textit{bjn}) & 4,000,000 * \\
South Sulawesi & Buginese (\textit{bug}) & 4,300,000 * \\
NTT & Uab Meto (\textit{aoz}) & 720,000 ** \\
West Papua & Wamesa (\textit{wad}) & 5,000 *** \\
- & Indonesia (ind) & 171,000,000 * \\
\bottomrule
\end{tabular}%
}
\caption{Provinces, languages (along with their ISO-639 codes), estimated speaker populations, and language status covered in \indoculturedialoguetitle, based on~\citep{ritchie-etal-2024-linguameta}. *Institutional language; **Stable language; ***Endangered language according to~\href{https://www.ethnologue.com/language/}{https://www.ethnologue.com/language/}, accessed on 11 May 2026.}
\label{tab:appendix-language-covered}
\end{table}

\section{MCQ Evaluation Details}
\label{sec:appendix-mcq-evaluation-details}

\subsection{MCQ Scores}
\label{sec:appendix-mcq-evaluation-details-a}
The more detailed dialogue-based MCQ cultural commonsense results are presented in Table~\ref{tab:mcq-main-result-appendix}, while scores across provinces and topics are in Table~\ref{tab:appendix-mcq-indo-province-breakdown},~\ref{tab:appendix-mcq-local-province-breakdown},~\ref{tab:appendix-mcq-indo-topic-breakdown}, and~\ref{tab:appendix-mcq-local-topic-breakdown}. Overall, Gemini-2.5-flash achieves the best performance on this benchmark, delivering consistent results across both settings, followed closely by other proprietary models, such as GPT-5.1 and Cohere-Command-A, which also demonstrate robust performance across all settings, including both province-specific and non-province-specific dialogue metrics.

Generic multilingual models lag substantially behind proprietary models, achieving average scores of 69.45 on Indonesian dialogues and 56.64 on local-language dialogues. In contrast, SEA-centric models consistently outperform generic multilingual baselines. Among them, Sahabat-AI-v1 Instruct achieves the strongest performance on Indonesian dialogues, trailing only the lowest-performing proprietary model (Cohere Command-A) by an average of 3.72 points.

We also observe that providing context improves performance across open-source model groups. For example, in the Indonesian dialogue setting, the average PS score for generic multilingual models increases by 2.17 with province context, whereas the \textasciitilde PS score improves by only 0.89.

\subsection{Effect of Pretraining in SEA Contexts}
\label{sec:appendix-mcq-evaluation-details-b}

\begin{figure}[t]
  \centering
  \includegraphics[width=0.9\columnwidth]{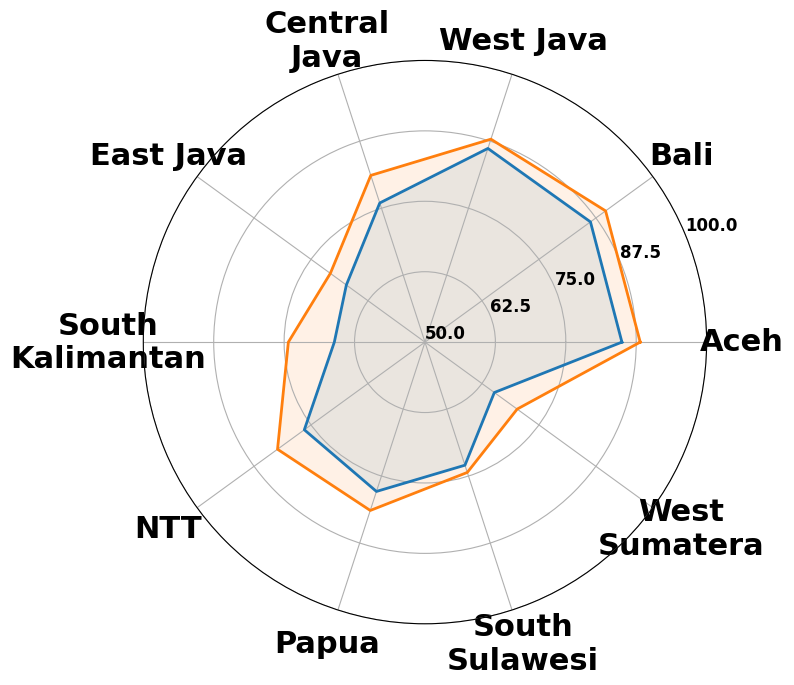}
  \caption{MCQ cultural commonsense evaluation score differences by province between \textcolor{blue}{\textit{Gemma2-9B Instruct}} (generic-multilingual model) and \textcolor{orange}{\textit{Sahabat-AI-v1 Instruct}} (SEA-centric model).}
  \label{fig:mcq-multi-vs-sea}
\end{figure}

We further examine the effect of further pretraining the multilingual model in Indonesian-related context in Indonesian (see Figure~\ref{fig:mcq-multi-vs-sea}, Table~\ref{tab:appendix-mcq-indo-province-breakdown}, and Table~\ref{tab:appendix-mcq-indo-topic-breakdown}) and local language settings (see Table~\ref{tab:appendix-mcq-local-province-breakdown} and Table~\ref{tab:appendix-mcq-local-topic-breakdown}) by comparing the best-performing generic multilingual and SEA-centric models. The SEA-centric model shows improvements across all provinces and topics, even for languages not explicitly included in its training data.

\setlength{\tabcolsep}{0.4em}
\setlength{\arrayrulewidth}{1.1pt}
\begin{table*}[bt]
\centering
\footnotesize
\begin{tabular}{l  ccc | ccc | ccc }
\toprule
\multirow{2}{*}{\textbf{Model}} 
& \multicolumn{3}{c|}{\textbf{Context: None}} 
& \multicolumn{3}{c|}{\textbf{Context: Prov.}} 
& \multicolumn{3}{c}{\textbf{Context: Prov. + Lang.}} \\
\cline{2-10}
& \textbf{Avg} & \textbf{PS} & \textbf{\textasciitilde PS}
& \textbf{Avg} & \textbf{PS} & \textbf{\textasciitilde PS}
& \textbf{Avg} & \textbf{PS} & \textbf{\textasciitilde PS} \\
\midrule
\midrule
\multicolumn{10}{l}{\yellowcell \textit{Indonesian Dialogue}} \\
\midrule

\multicolumn{10}{c}{\textbf{Proprietary Models}} \\
GPT-5.1 
& 89.23\textsubscript{(2)} & 86.44 & 94.72 
& 89.86\textsubscript{(2)} & 87.18 & 95.12 
& 89.95\textsubscript{(2)} & 87.38 & 94.99 \\

Gemini-2.5-flash
& \textbf{91.46}\textsubscript{(1)} & \textbf{89.66} & \textbf{94.99} 
& \textbf{91.55}\textsubscript{(1)} & \textbf{89.80} & \textbf{94.99} 
& \textbf{91.68}\textsubscript{(1)} & \textbf{89.99} & \textbf{94.99} \\

Cohere-Command-A
& 84.74\textsubscript{(3)} & 80.74 & 92.61 
& 85.54\textsubscript{(3)} & 82.28 & 91.95 
& 85.01\textsubscript{(3)} & 81.41 & 92.08 \\

\hline
\textit{average proprietary} 
& \textit{88.48} & \textit{85.61} & \textit{94.11}
& \textit{88.98} & \textit{86.42} & \textit{94.02}
& \textit{88.88} & \textit{86.26} & \textit{94.02} \\

\midrule

\multicolumn{10}{c}{\textbf{Multilingual Models}} \\
Qwen3-8B 
& 68.95\textsubscript{(7)} & 66.64 & 73.84 
& 70.82\textsubscript{(7)} & 68.52 & 75.33 
& 70.28\textsubscript{(7)} & 67.99 & 74.80 \\

Llama3.1-8B Instruct 
& 62.63\textsubscript{(8)} & 60.13 & 67.55 
& 65.44\textsubscript{(8)} & 63.62 & 69.00 
& 65.44\textsubscript{(8)} & 63.89 & 68.74 \\

Gemma2-9B Instruct 
& \bold{76.78}\textsubscript{(6)} & \bold{72.82} & \bold{84.56} 
& \bold{77.45}\textsubscript{(6)} & \bold{73.96} & \bold{84.30} 
& \bold{77.71}\textsubscript{(6)} & \bold{74.09} & \bold{84.83} \\

\hline
\textit{average multilingual}
& \textit{69.45} & \textit{66.53} & \textit{75.32}
& \textit{71.24} & \textit{68.70} & \textit{76.21}
& \textit{71.14} & \textit{68.66} & \textit{76.12} \\
\midrule

\multicolumn{10}{c}{\textbf{SEA-Centric Models}} \\
Sailor2-8B Chat 
& 80.07\textsubscript{(5)} & 76.44 & 87.20 
& 81.01\textsubscript{(5)} & 77.45 & \bold{87.99} 
& 81.32\textsubscript{(5)} & 77.79 & \bold{88.26} \\

SEA-LION-v3.5-8B 
& 36.57\textsubscript{(9)} & 36.85 & 36.02 
& 36.57\textsubscript{(9)} & 36.85 & 36.02
& 36.57\textsubscript{(9)} & 36.85 & 36.02 \\

Sahabat-AI-v1 Instruct 
& \bold{80.87}\textsubscript{(4)} & \bold{77.25} & \bold{87.99} 
& \bold{81.72}\textsubscript{(4)} & \bold{78.99} & 87.07 
& \bold{81.54}\textsubscript{(4)} & \bold{78.52} & 87.47 \\

\hline
\textit{average SEA-centric*}
& \textit{80.47} & \textit{76.85} & \textit{87.60}
& \textit{81.37} & \textit{78.22} & \textit{87.53}
& \textit{81.43} & \textit{78.16} & \textit{87.87} \\

\midrule
\midrule
\multicolumn{10}{l}{\yellowcell \textit{Local Indonesian Dialogue}} \\
\midrule

\multicolumn{10}{c}{\textbf{Proprietary Models}} \\
GPT-5.1 
& 82.31\textsubscript{(2)} & 79.07 & 88.67 
& 82.52\textsubscript{(2)} & 78.86 & 89.71 
& 82.70\textsubscript{(2)} & 79.26 & 89.45 \\
Gemini-2.5-flash
& \textbf{87.77}\textsubscript{(1)} & \textbf{85.23} & \textbf{92.74} 
& \textbf{87.90}\textsubscript{(1)} & \textbf{85.70} & \textbf{92.22} 
& \textbf{87.72}\textsubscript{(1)} & \textbf{85.23} & \textbf{92.61} \\
Cohere-Command-A
& 75.09\textsubscript{(3)} & 70.07 & 84.96 
& 75.67\textsubscript{(3)} & 71.21 & 84.43 
& 74.64\textsubscript{(3)} & 70.00 & 83.77 \\
\hline
\textit{average proprietary} 
& \textit{81.72} & \textit{78.12} & \textit{88.79}
& \textit{82.03} & \textit{78.59} & \textit{88.79}
& \textit{81.69} & \textit{78.16} & \textit{88.61} \\

\midrule

\multicolumn{10}{c}{\textbf{Multilingual Models}} \\
Qwen3-8B 
& 52.89\textsubscript{(7)} & 48.86 & 60.82 
& 55.29\textsubscript{(7)} & 51.28 & 63.19 
& 54.72\textsubscript{(7)} & 50.74 & 62.53 \\
Llama3.1-8B Instruct 
& 51.87\textsubscript{(8)} & 49.60 & 56.33 
& 54.09\textsubscript{(8)} & 52.55 & 57.12 
& 53.87\textsubscript{(8)} & 51.95 & 57.65 \\
Gemma2-9B Instruct 
& \bold{65.17}\textsubscript{(6)} & \bold{59.80} & \bold{75.73} 
& \bold{66.01}\textsubscript{(6)} & \bold{61.21} & \bold{75.46} 
& \bold{65.88}\textsubscript{(6)} & \bold{60.87} & \bold{75.73} \\
\hline
\textit{average multilingual}
& \textit{56.64} & \textit{52.75} & \textit{64.29}
& \textit{58.46} & \textit{55.01} & \textit{65.26}
& \textit{58.15} & \textit{54.52} & \textit{65.30} \\
\midrule

\multicolumn{10}{c}{\textbf{SEA-Centric Models}} \\
Sailor2-8B Chat 
& \bold{70.15}\textsubscript{(4)} & \bold{64.50} & \bold{81.27} 
& \bold{71.44}\textsubscript{(4)} & \bold{66.24} & \bold{81.66} 
& \bold{72.02}\textsubscript{(4)} & \bold{66.85} & \bold{82.19} \\
SEA-LION-v3.5-8B-R 
& 36.57\textsubscript{(9)} & 36.85 & 36.02 
& 36.57\textsubscript{(9)} & 36.85 & 36.02 
& 36.57\textsubscript{(9)} & 36.85 & 36.02 \\
Sahabat-AI-v1 Instruct 
& 69.08\textsubscript{(5)} & 63.62 & 79.82 
& 69.80\textsubscript{(5)} & 64.77 & 79.68 
& 69.93\textsubscript{(5)} & 65.17 & 79.29 \\
\hline
\textit{average SEA-centric*}
& \textit{69.62} & \textit{63.56} & \textit{80.55}
& \textit{70.62} & \textit{65.51} & \textit{80.67}
& \textit{70.98} & \textit{66.01} & \textit{80.71} \\
\bottomrule
\end{tabular}
\caption{Model performance in \textbf{dialogue-based cultural commonsense MCQ} under three context settings: no context, province context, and province plus language context. Scores are reported as average (Avg), province-specific (PS), and non-province-specific ($\sim$PS). (*) The SEA-LION-v3.5-8B-R scores are excluded.}
\label{tab:mcq-main-result-appendix}
\end{table*}

\setlength{\tabcolsep}{0.35em}
\begin{table}[t]
\centering
\small
\begin{tabular}{l l ccc}
\toprule
\textbf{Model} & \textbf{Province} & \textbf{Avg} & \textbf{CS} & \textbf{$\sim$CS} \\
\midrule
& Aceh        & 92.68 & 90.80 & 97.22 \\
& Bali        & 95.44 & 95.65 & 94.74 \\
& W. Java    & \greencell96.10 & \greencell97.01 & 94.85 \\
& C. Java   & 92.27 & 91.36 & 94.37 \\
Gemini-2.5 & E. Java    & \redcell87.72 & \redcell83.33 & 96.49 \\
Flash & S. Kalimantan  & 88.41 & 87.69 & \redcell92.11 \\
& NTT         & \greencell100.00 & \greencell100.00 & \greencell100.00 \\
& W. Papua       & 92.89 & 91.93 & \greencell100.00 \\
& S. Sulawesi    & 91.89 & 91.04 & \greencell100.00 \\
& W. Sumatera    & \redcell81.94 & \redcell79.82 & \redcell88.16 \\
& Indonesia   & 96.00 & -- & -- \\
\hline
 & Aceh        & 84.96 & 81.61 & \greencell93.06 \\
 & Bali        & \greencell86.31 & \greencell84.78 & 91.23 \\
 & W. Java    & \greencell86.15 & \greencell84.33 & 88.66 \\
 & C. Java   & 75.97 & 72.84 & 83.10 \\
Gemma-2 9B & E. Java    & 67.25 & \redcell58.77 & 84.21 \\
Instruct & S. Kalimantan  & \redcell66.09 & 63.08 & 81.58 \\
 & NTT         & 76.47 & 78.57 & \redcell66.67 \\
 & W. Papua       & 77.87 & 75.78 & \greencell93.33 \\
 & S. Sulawesi    & 72.97 & 73.13 & \redcell71.43 \\
 & W. Sumatera    & \redcell65.22 & \redcell61.43 & 76.32 \\
 & Indonesia   & 82.00 & -- & -- \\
\hline
 & Aceh        & \greencell88.21 & 84.48 & \greencell97.22 \\
 & Bali        & \greencell89.63 & \greencell88.04 & 94.74 \\
 & W. Java    & 87.88 & \greencell88.81 & 86.60 \\
 & C. Java   & 81.12 & 78.40 & 87.32 \\
 Sahabat-AI-v1 & E. Java    & \redcell70.76 & \redcell59.65 & 92.98 \\
 Instruct & S. Kalimantan  & 74.25 & 72.31 & 84.21 \\
 & NTT         & 82.35 & 78.57 & \greencell100.00 \\
 & W. Papua       & 81.42 & 80.27 & 90.00 \\
 & S. Sulawesi    & 74.32 & 74.63 & \redcell71.43 \\
 & W. Sumatera    & \redcell70.23 & \redcell65.92 & \redcell82.89 \\
 & Indonesia   & 85.60 & -- & -- \\
\bottomrule
\end{tabular}
\caption{Performance of dialogue-based cultural commonsense MCQ in Indonesian by model across provinces. Scores reported as average (Avg), country-specific (CS), and non-country-specific ($\sim$CS). W, C, E, and S denote west, central, east, and south, respectively.}
\label{tab:appendix-mcq-indo-province-breakdown}
\end{table}

\setlength{\tabcolsep}{0.35em}
\begin{table}[t]
\centering
\small
\begin{tabular}{l l ccc}
\toprule
\textbf{Model} & \textbf{Province} & \textbf{Avg} & \textbf{CS} & \textbf{$\sim$CS} \\
\midrule
& Aceh        & 91.46 & 90.80 & 93.06 \\
& Bali        & \greencell95.44 & \greencell96.20 & 92.98 \\
& W. Java     & 93.94 & \greencell94.78 & 92.78 \\
& C. Java     & \greencell94.85 & 93.83 & \greencell97.18 \\
Gemini-2.5 & E. Java    & 87.13 & 83.33 & \greencell94.74 \\
Flash & S. Kalimantan  & 85.41 & 83.59 & \greencell94.74 \\
& NTT         & \redcell35.29 & \redcell35.71 & \redcell33.33 \\
& W. Papua       & \redcell68.77 & \redcell69.06 & \redcell66.67 \\
& S. Sulawesi    & 83.78 & 85.07 & 71.43 \\
& W. Sumatera    & 83.28 & 81.61 & 88.16 \\
& Indonesia   & 96.00 & -- & -- \\
\hline
 & Aceh        & 67.89 & 64.37 & 76.39 \\
 & Bali        & \greencell72.61 & 69.02 & \greencell84.21 \\
 & W. Java     & \greencell79.65 & \greencell78.36 & \greencell81.44 \\
 & C. Java     & 71.67 & \greencell74.07 & 66.20 \\
Gemma-2 9B & E. Java    & 59.65 & 52.63 & 73.68 \\
Instruct & S. Kalimantan  & 61.37 & 58.46 & 76.32 \\
 & NTT         & \redcell29.41 & \redcell28.57 & \redcell33.33 \\
 & W. Papua       & 50.99 & 50.22 & 56.67 \\
 & S. Sulawesi    & \redcell37.84 & \redcell38.81 & \redcell28.57 \\
 & W. Sumatera    & 53.51 & 49.78 & 64.47 \\
 & Indonesia   & 82.00 & -- & -- \\
\hline
 & Aceh        & 69.51 & 64.94 & 80.56 \\
 & Bali        & \greencell79.67 & \greencell80.43 & 77.19 \\
 & W. Java     & \greencell83.12 & \greencell82.84 & \greencell83.51 \\
 & C. Java     & 76.39 & 73.46 & 83.10 \\
 Sailor2-8B & E. Java    & 71.35 & 63.16 & \greencell87.72 \\
 Chat & S. Kalimantan  & 64.81 & 61.54 & 81.58 \\
 & NTT         & 58.82 & 64.29 & \redcell33.33 \\
 & W. Papua       & \redcell51.38 & \redcell50.22 & 60.00 \\
 & S. Sulawesi    & \redcell41.89 & \redcell41.79 & \redcell42.86 \\
 & W. Sumatera    & 61.20 & 57.85 & 71.05 \\
 & Indonesia   & 86.80 & -- & -- \\
\bottomrule
\end{tabular}
\caption{Performance of dialogue-based cultural commonsense MCQ in Indonesian local languages by model across provinces. Scores reported as average (Avg), country-specific (CS), and non-country-specific ($\sim$CS). W, C, E, and S denote west, central, east, and south, respectively.}
\label{tab:appendix-mcq-local-province-breakdown}
\end{table}

\setlength{\tabcolsep}{0.35em}
\begin{table}[t]
\centering
\small
\begin{tabular}{l l ccc}
\toprule
\textbf{Model} & \textbf{Topic} & \textbf{Avg} & \textbf{CS} & \textbf{$\sim$CS} \\
\midrule
& Food              & 91.95 & 92.28 & \redcell90.20 \\
& Wedding           & \redcell86.97 & \redcell85.92 & 91.67 \\
& Family    & \redcell87.79 & 86.99 & \redcell89.80 \\
& Birth & 90.09 & 87.12 & 95.00 \\
& Death             & 89.15 & 86.42 & 93.75 \\
Gemini-2.5 & Religion     & \greencell97.33 & \greencell96.77 & 98.25 \\
Flash & Agriculture            & \greencell94.81 & \greencell94.34 & 95.83 \\
& Farm      & 88.24 & \redcell84.13 & \greencell100.00 \\
& Art                 & 91.98 & 91.46 & \greencell100.00 \\
& Game            & 92.41 & 92.54 & 91.67 \\
& Daily          & 93.50 & 93.15 & 94.00 \\
& Socio-religious      & 90.29 & 86.78 & 98.15 \\
\hline
& Food              & 73.07 & 72.79 & \redcell74.51 \\
& Wedding           & \redcell70.88 & \redcell67.61 & 85.42 \\
& Family    & 76.16 & 72.36 & 85.71 \\
& Birth & 78.77 & 72.73 & 88.75 \\
& Death             & 76.74 & 70.37 & 87.50 \\
Gemma-2 9B & Religion     & \greencell84.00 & \greencell87.10 & 78.95 \\
Instruct & Agriculture            & \greencell88.31 & \greencell84.91 & \greencell95.83 \\
& Farm      & \redcell70.59 & \redcell65.08 & 86.36 \\
& Art                 & 74.53 & 73.37 & \greencell92.31 \\
& Game            & 73.42 & 73.13 & \redcell75.00 \\
& Daily          & 80.49 & 72.60 & 92.00 \\
& Socio-religious      & 76.57 & 71.07 & 88.89 \\
\hline
& Food              & 78.64 & 77.94 & \redcell82.35 \\
& Wedding           & 76.63 & 74.18 & 87.50 \\
& Family    & 81.98 & 78.86 & 89.80 \\
& Birth & 80.19 & 74.24 & 90.00 \\
& Death             & 80.62 & 76.54 & 87.50 \\
Sahabat-AI-v1 & Religion     & \greencell86.00 & \greencell86.02 & 85.96 \\
Instruct & Agriculture            & \greencell88.31 & \greencell84.91 & \greencell95.83 \\
& Farm      & \redcell76.47 & \redcell71.43 & 90.91 \\
& Art                 & 78.30 & 77.39 & 92.31 \\
& Game            & \redcell73.42 & \redcell73.13 & \redcell75.00 \\
& Daily          & 84.55 & 78.08 & 94.00 \\
& Socio-religious      & 82.86 & 77.69 & \greencell94.44 \\
\bottomrule
\end{tabular}
\caption{Performance of dialogue-based cultural commonsense MCQ in Indonesian by model across topics. Scores reported as average (Avg), country-specific (CS), and non-country-specific ($\sim$CS).}
\label{tab:appendix-mcq-indo-topic-breakdown}
\end{table}

\setlength{\tabcolsep}{0.35em}
\begin{table}[t]
\centering
\small
\begin{tabular}{l l ccc}
\toprule
\textbf{Model} & \textbf{Topic} & \textbf{Avg} & \textbf{CS} & \textbf{$\sim$CS} \\
\midrule
& Food              & \redcell84.21 & 84.19 & \redcell84.31 \\
& Wedding           & \redcell82.76 & 82.16 & \redcell85.42 \\
& Family            & 84.88 & 82.11 & 91.84 \\
& Birth             & 87.26 & 83.33 & 93.75 \\
& Death             & 85.27 & 83.95 & 87.50 \\
Gemini-2.5 & Religion     & \greencell92.67 & \greencell92.47 & 92.98 \\
Flash & Agriculture       & 90.91 & 90.57 & 91.67 \\
& Farm                 & 83.53 & \redcell80.95 & 90.91 \\
& Art                  & \greencell92.45 & \greencell91.96 & \greencell100.00 \\
& Game                 & 86.08 & 85.07 & 91.67 \\
& Daily                & 89.43 & 87.67 & 92.00 \\
& Socio-religious      & 85.14 & \redcell80.99 & \greencell94.44 \\
\hline
& Food              & \redcell52.32 & \redcell51.47 & \redcell56.86 \\
& Wedding           & 64.75 & 61.97 & 77.08 \\
& Family            & 60.47 & 54.47 & 75.51 \\
& Birth             & 62.74 & 56.06 & 73.75 \\
& Death             & 67.44 & 62.96 & 75.00 \\
Gemma-2 9B & Religion     & \greencell74.00 & \greencell74.19 & 73.68 \\
Instruct & Agriculture    & \greencell72.73 & \greencell75.47 & 66.67 \\
& Farm                 & 57.65 & \redcell50.79 & \greencell77.27 \\
& Art                  & 64.62 & 62.81 & \greencell92.31 \\
& Game                 & \redcell55.70 & 55.22 & \redcell58.33 \\
& Daily                & 67.48 & 63.01 & 74.00 \\
& Socio-religious      & 67.43 & 64.46 & 74.07 \\
\hline
& Food              & \redcell63.16 & 62.87 & \redcell64.71 \\
& Wedding           & \redcell64.37 & 61.97 & 75.00 \\
& Family            & 66.28 & 60.98 & 79.59 \\
& Birth             & 66.98 & \redcell58.33 & 81.25 \\
& Death             & 66.67 & 65.43 & \redcell68.75 \\
Sailor2-8B & Religion     & 77.33 & 72.04 & \greencell85.96 \\
Chat & Agriculture       & \greencell80.52 & \greencell79.25 & 83.33 \\
& Farm                 & 67.06 & 61.90 & 81.82 \\
& Art                  & \greencell74.06 & \greencell73.37 & 84.62 \\
& Game                 & 67.09 & 64.18 & 83.33 \\
& Daily                & 66.67 & \redcell58.90 & 78.00 \\
& Socio-religious      & 68.00 & 60.33 & \greencell85.19 \\
\bottomrule
\end{tabular}
\caption{Performance of dialogue-based cultural commonsense MCQ in Indonesian local languages by model across topics. Scores reported as average (Avg), country-specific (CS), and non-country-specific ($\sim$CS).}
\label{tab:appendix-mcq-local-topic-breakdown}
\end{table}

\section{MT Evaluation Details}
\label{sec:appendix-mt-evaluation-details}

We present the full machine translation results in Table~\ref{tab:mt-main-result-appendix}. Overall, models perform substantially better when translating from local Indonesian languages into Indonesian than in the reverse direction. This likely reflects the stronger representation and preservation of Indonesian in current language models, while many local Indonesian languages remain under-resourced. In the local-to-Indonesian setting, the best-performing model (GPT-5.1) achieves an overall score of 4.69 and a BLEU score of 50.54.

Consistent with the main results, generic multilingual models perform considerably worse than proprietary and SEA-centric models. For example, the strongest multilingual baseline, Gemma-2-9B Instruct, achieves only 20.54 BLEU and a 2.73 overall score in the Indonesian-to-local setting. SEA-centric models generally perform better, with Sailor2-8B Chat approaching Cohere-Command-A in the Indonesian-to-local direction, while Sahabat-AI-v1 Instruct achieves competitive performance in the local-to-Indonesian setting with an overall score of 3.98.

Supervised fine-tuning further improves performance for both Gemma-2-9B Instruct and Sahabat-AI-v1 Instruct across both translation directions, with especially large gains in local-to-Indonesian translation. We additionally report province-level translation results in Table~\ref{tab:mt-indo-local-by-region-0shot},~\ref{tab:mt-indo-local-by-region-sft},~\ref{tab:mt-local-indo-by-region-0shot}, and~\ref{tab:mt-local-indo-by-region-sft}. Across most models, Javanese tends to achieve stronger performance than other local languages, likely due to its relatively larger resource availability and speaker population.

\setlength{\tabcolsep}{0.35em}
\begin{table*}[t]
\centering
\footnotesize
\begin{tabular}{l c c ccccc}
\toprule
\multirow{2}{*}{\textbf{Model}}
& \textbf{BLEU Scores} 
& \textbf{BERTScore} 
& \multicolumn{5}{c}{\textbf{LLM as a Judge (1--5)}} \\
\cmidrule(lr){2-2} \cmidrule(lr){3-3} \cmidrule(lr){4-8}
& \textbf{B4} & \textbf{F1}
& \textbf{adequacy} & \textbf{fluency} & \textbf{register} & \textbf{terminology} & \textbf{overall} \\
\midrule
\midrule
\multicolumn{8}{l}{\purplecell \textit{Indonesian $\rightarrow$ local Indonesian languages}} \\
\midrule
\multicolumn{8}{c}{\textbf{Proprietary Models (0-shot)}} \\
GPT-5.1 
& 27.53 & 84.02 & 4.35 & 3.80 & 3.75 & 4.33 & 3.87 \textsubscript{(2)} \\
Gemini-2.5-flash 
& \textbf{28.36} & \textbf{84.33} & \textbf{4.48} & \textbf{3.91} & \textbf{3.91} & \textbf{4.48} & \textbf{4.00} \textsubscript{(1)} \\
Cohere-Command-A 
& 21.45 & 81.40 & 3.35 & 2.82 & 2.69 & 3.29 & 2.93 \textsubscript{(5)} \\
\midrule

\multicolumn{8}{c}{\textbf{Multilingual Models (0-shot)}} \\
Qwen3-8B 
& 19.78 \textsubscript{($\pm$0.1)} & 78.74 \textsubscript{($\pm$0.1)} & 2.82 & 2.11 & 1.90 & 2.73 & 2.22 \textsubscript{(11)} \\
Llama-3.1-8B Instruct 
& 17.62 \textsubscript{($\pm$0.1)} & 78.71 \textsubscript{($\pm$0.1)} & 2.79 & 2.27 & 2.21 & 2.81 & 2.49 \textsubscript{(10)} \\
Gemma-2-9b Instruct 
& \textbf{20.54} \textsubscript{($\pm$0.1)} & \textbf{80.15} \textsubscript{($\pm$0.1)} & \textbf{3.20} & \textbf{2.49} & \textbf{2.40} & \textbf{3.11} & \textbf{2.73} \textsubscript{(7)} \\
\midrule

\multicolumn{8}{c}{\textbf{SEA-Centric Models (0-shot)}} \\
Sailor2-8B Chat 
& 15.44 \textsubscript{($\pm$0.1)} & 77.58 \textsubscript{($\pm$0.1)} & 3.38 & \textbf{2.85} & \textbf{2.65} & \textbf{3.42} & \textbf{2.94} \textsubscript{(4)} \\
SEA-LION-v3.5-8B 
& 19.34 \textsubscript{($\pm$0.1)} & 78.98 \textsubscript{($\pm$0.1)} & 3.18 & 2.61 & 2.44 & 3.11 & 2.72 \textsubscript{(9)} \\
Sahabat-AI-v1 Instruct 
& \textbf{21.70} \textsubscript{($\pm$0.0)} & \textbf{80.51} \textsubscript{($\pm$0.0)} & \textbf{3.57} & 2.63 & 2.46 & 3.21 & 2.80 \textsubscript{(6)} \\
\midrule

\multicolumn{8}{c}{\textbf{SFT Models}} \\
Gemma-2-9b Instruct 
& \green{$\Uparrow$} 22.60 \textsubscript{($\pm$0.0)} & \red{$\Downarrow$} 79.39 \textsubscript{($\pm$0.0)} & \red{$\Downarrow$} 3.16 & \green{$\Uparrow$} 2.54 & \green{$\Uparrow$} 2.46 & \green{$\Uparrow$} 3.14 & \textcolor{gray}{$\approx$} 2.73 \textsubscript{(7)} \\
Sahabat-AI-v1 Instruct 
& \green{$\Uparrow$} \textbf{24.56} \textsubscript{($\pm$0.0)} & \red{$\Downarrow$} \textbf{80.30} \textsubscript{($\pm$0.0)} & \red{$\Downarrow$} \textbf{3.53} & \green{$\Uparrow$} \textbf{3.03} & \green{$\Uparrow$} \textbf{2.97} & \green{$\Uparrow$} \textbf{3.55} & \green{$\Uparrow$} \textbf{3.14} \textsubscript{(3)} \\
\midrule
\midrule
\multicolumn{8}{l}{\purplecell \textit{local Indonesian $\rightarrow$ Indonesian languages}} \\
\midrule
\multicolumn{8}{c}{\textbf{Proprietary Models (0-shot)}} \\
GPT-5.1 
& 50.54 & 90.78 & \textbf{4.64} & \textbf{4.96} & \textbf{4.88} & \textbf{4.70} & \textbf{4.69} \textsubscript{(1)} \\
Gemini-3-flash-preview 
& \textbf{52.74} & \textbf{91.46} & 4.60 & \textbf{4.96} & 4.85 & 4.68 & 4.67 \textsubscript{(2)} \\
Cohere-Command-A 
& 42.00 & 88.41 & 3.96 & 4.85 & 4.52 & 4.17 & 4.17 \textsubscript{(4)} \\
\midrule

\multicolumn{8}{c}{\textbf{Multilingual Models (0-shot)}} \\
Qwen3-8B 
& 24.75 \textsubscript{($\pm$0.3)} & 80.25 \textsubscript{($\pm$0.2)} & 2.53 & 2.80 & 2.70 & 2.65 & 2.54 \textsubscript{(11)} \\
Llama-3.1-8B Instruct 
& 26.60 \textsubscript{($\pm$0.2)} & 83.71 \textsubscript{($\pm$0.1)} & 2.66 & 3.61 & 3.19 & 2.93 & 2.89 \textsubscript{(10)} \\
Gemma-2-9b Instruct 
& \textbf{30.41} \textsubscript{($\pm$0.1)} & \textbf{84.46} \textsubscript{($\pm$0.1)} & \textbf{3.30} & \textbf{4.14} & \textbf{3.74} & \textbf{3.49} & \textbf{3.47} \textsubscript{(7)} \\
\midrule

\multicolumn{8}{c}{\textbf{SEA-Centric Models (0-shot)}} \\
Sailor2-8B Chat 
& 22.84 \textsubscript{($\pm$0.1)} & 80.89 \textsubscript{($\pm$0.1)} & 3.25 & 4.14 & 3.68 & 3.41 & 3.40 \textsubscript{(8)} \\
SEA-LION-v3.5-8B 
& 31.66 \textsubscript{($\pm$0.0)} & 84.87 \textsubscript{($\pm$0.0)} & 3.13 & 4.00 & 3.58 & 3.31 & 3.34 \textsubscript{(9)} \\
Sahabat-AI-v1 Instruct 
& \textbf{39.20} \textsubscript{($\pm$0.0)} & \textbf{87.41} \textsubscript{($\pm$0.0)} & \textbf{3.80} & \textbf{4.61} & \textbf{4.28} & \textbf{3.96} & \textbf{3.98} \textsubscript{(5)} \\
\midrule

\multicolumn{8}{c}{\textbf{SFT Models}} \\
Gemma-2-9b Instruct 
& \green{$\Uparrow$} 45.79 \textsubscript{($\pm$0.0)} & \green{$\Uparrow$} 89.45 \textsubscript{($\pm$0.1)} & \green{$\Uparrow$} 3.69 & \green{$\Uparrow$} 4.81 & \green{$\Uparrow$} 4.31 & \green{$\Uparrow$} 3.87 & \green{$\Uparrow$} 3.93 \textsubscript{(6)} \\
Sahabat-AI-v1 Instruct 
& \green{$\Uparrow$} \textbf{49.81} \textsubscript{($\pm$0.0)} & \green{$\Uparrow$} \textbf{90.39} \textsubscript{($\pm$0.0)} & \green{$\Uparrow$} \textbf{3.99} & \green{$\Uparrow$} \textbf{4.93} & \green{$\Uparrow$} \textbf{4.51} & \green{$\Uparrow$} \textbf{4.14} & \green{$\Uparrow$} \textbf{4.20} \textsubscript{(3)} \\
\bottomrule

\end{tabular}
\caption{Machine translation results for \textbf{Indonesian $\rightarrow$ local} (top) and \textbf{local $\rightarrow$ Indonesian} (bottom) under the \textbf{Province + Language} context. We report BLEU-4, BERTScore F1, and LLM-as-a-judge scores for adequacy, fluency, register, terminology, and overall quality.}
\label{tab:mt-main-result-appendix}
\end{table*}

\setlength{\tabcolsep}{0.35em}
\begin{table*}[t]
\centering
\small
\begin{tabular}{l cccc ccc ccccc}
\toprule
\multirow{2}{*}{\textbf{Province}}
& \multicolumn{4}{c}{\textbf{BLEU Scores}}
& \multicolumn{3}{c}{\textbf{BERTScore}}
& \multicolumn{5}{c}{\textbf{LLM as a judge (1--5)}} \\
\cmidrule(lr){2-5} \cmidrule(lr){6-8} \cmidrule(lr){9-13}
& \textbf{B1} & \textbf{B2} & \textbf{B3} & \textbf{B4}
& \textbf{P} & \textbf{R} & \textbf{F1}
& \textbf{adequacy} & \textbf{fluency} & \textbf{register} & \textbf{terminology} & \textbf{overall} \\
\midrule

\multicolumn{13}{c}{\textbf{Gemini-3-Flash (reasoning none)}} \\
Aceh 
& 0.39 & 0.25 & 0.16 & 0.09 & 0.78 & 0.80 & 0.79 & 4.67 & 4.10 & 4.26 & 4.68 & 4.26 \\
Bali 
& 0.32 & 0.19 & 0.11 & 0.07 & 0.80 & 0.81 & 0.80 & 4.61 & 3.99 & 3.99 & 4.67 & 4.02 \\
West Java 
& 0.51 & 0.38 & 0.28 & 0.19 & 0.84 & 0.85 & 0.84 & 4.78 & 3.97 & 3.70 & 4.75 & 3.99 \\
Central Java 
& 0.59 & 0.47 & 0.38 & 0.29 & 0.87 & 0.87 & 0.87 & \textbf{4.86} & \textbf{4.29} & 4.11 & \textbf{4.83} & \textbf{4.30} \\
East Java 
& 0.57 & 0.44 & 0.33 & 0.25 & 0.86 & 0.86 & 0.86 & 4.80 & 4.13 & 4.03 & 4.77 & 4.15 \\
South Kalimantan
& 0.62 & 0.49 & 0.39 & 0.30 & 0.87 & 0.88 & 0.87 & 4.74 & 4.09 & 4.13 & 4.76 & 4.15 \\
NTT 
& 0.20 & 0.09 & 0.04 & 0.02 & 0.74 & 0.74 & 0.74 & 4.00 & 3.00 & 3.40 & 3.80 & 3.47 \\
Papua 
& 0.14 & 0.07 & 0.03 & 0.01 & 0.69 & 0.69 & 0.69 & 2.32 & 1.87 & 1.96 & 2.32 & 2.18 \\
South Sulawesi
& 0.27 & 0.16 & 0.08 & 0.04 & 0.78 & 0.77 & 0.78 & 4.25 & 3.61 & 3.73 & 4.21 & 3.76 \\
West Sumatera
& \textbf{0.68} & \textbf{0.57} & \textbf{0.48} & \textbf{0.40} & \textbf{0.89} & \textbf{0.90} & \textbf{0.90} & 4.80 & 4.07 & \textbf{4.14} & 4.80 & 4.14 \\

\midrule
\multicolumn{13}{c}{\textbf{Gemma-2-9B Instruct}} \\
Aceh 
& 0.33 & 0.21 & 0.12 & 0.07 & 0.77 & 0.76 & 0.77 & 2.62 & 1.89 & 1.89 & 2.47 & 2.30 \\
Bali 
& 0.21 & 0.11 & 0.06 & 0.03 & 0.75 & 0.74 & 0.75 & 2.52 & 1.87 & 1.90 & 2.55 & 2.33 \\
West Java 
& \textbf{0.50} & \textbf{0.37} & \textbf{0.26} & \textbf{0.18} & \textbf{0.84} & \textbf{0.83} & \textbf{0.84} & 3.49 & 2.53 & 2.41 & 3.26 & 2.85 \\
Central Java 
& 0.48 & 0.35 & 0.24 & 0.16 & 0.83 & 0.81 & 0.82 & 3.47 & 2.61 & 2.53 & 3.26 & 2.89 \\
East Java 
& 0.45 & 0.31 & 0.21 & 0.13 & 0.82 & 0.81 & 0.82 & 3.62 & 2.68 & 2.58 & 3.49 & 3.03 \\
South Kalimantan
& 0.48 & 0.34 & 0.24 & 0.16 & 0.82 & 0.80 & 0.81 & 3.49 & 2.07 & 1.66 & 3.20 & 2.30 \\
NTT 
& 0.09 & 0.04 & 0.02 & 0.01 & 0.65 & 0.64 & 0.64 & 1.27 & 1.07 & 1.07 & 1.40 & 1.13 \\
Papua 
& 0.15 & 0.07 & 0.03 & 0.02 & 0.66 & 0.65 & 0.65 & 1.39 & 1.12 & 1.07 & 1.37 & 1.18 \\
South Sulawesi
& 0.16 & 0.09 & 0.04 & 0.02 & 0.69 & 0.66 & 0.68 & 2.13 & 1.61 & 1.58 & 2.12 & 1.87 \\
West Sumatera
& 0.46 & 0.32 & 0.22 & 0.14 & 0.83 & 0.82 & 0.82 & \textbf{3.76} & \textbf{2.93} & \textbf{2.88} & \textbf{3.81} & \textbf{3.14} \\

\midrule
\multicolumn{13}{c}{\textbf{Sahabat-AI-v1 Instruct}} \\
Aceh 
& 0.34 & 0.22 & 0.14 & 0.08 & 0.77 & 0.75 & 0.76 & 2.80 & 1.26 & 1.04 & 1.88 & 1.55 \\
Bali 
& 0.22 & 0.12 & 0.07 & 0.03 & 0.75 & 0.73 & 0.74 & 3.18 & 1.83 & 1.66 & 2.60 & 2.33 \\
West Java 
& 0.53 & \textbf{0.40} & \textbf{0.30} & 0.21 & \textbf{0.85} & \textbf{0.84} & \textbf{0.85} & 4.57 & 3.69 & 3.52 & 4.49 & 3.73 \\
Central Java 
& \textbf{0.54} & \textbf{0.40} & \textbf{0.30} & \textbf{0.22} & \textbf{0.85} & 0.83 & 0.84 & 4.49 & 3.67 & 3.50 & 4.40 & 3.70 \\
East Java 
& 0.50 & 0.36 & 0.26 & 0.18 & 0.84 & 0.83 & 0.83 & \textbf{4.58} & \textbf{3.71} & \textbf{3.57} & \textbf{4.51} & \textbf{3.75} \\
South Kalimantan
& 0.45 & 0.31 & 0.21 & 0.14 & 0.82 & 0.81 & 0.81 & 3.49 & 1.64 & 1.28 & 2.37 & 1.99 \\
NTT 
& 0.08 & 0.05 & 0.02 & 0.01 & 0.66 & 0.65 & 0.66 & 1.67 & 1.20 & 1.00 & 1.60 & 1.27 \\
Papua 
& 0.17 & 0.09 & 0.04 & 0.02 & 0.68 & 0.65 & 0.66 & 1.29 & 1.09 & 1.01 & 1.18 & 1.09 \\
South Sulawesi 
& 0.17 & 0.10 & 0.04 & 0.02 & 0.69 & 0.65 & 0.67 & 2.00 & 1.24 & 1.09 & 1.67 & 1.33 \\
West Sumatera 
& 0.47 & 0.33 & 0.23 & 0.16 & 0.83 & 0.82 & 0.82 & 3.79 & 2.66 & 2.45 & 3.52 & 2.92 \\

\bottomrule
\end{tabular}
\caption{Translation results for \textbf{Indonesian $\rightarrow$ local} by provinces. We compare the best proprietary model (Gemini-3-flash-preview), best multilingual model (Gemma-2-9b-it), and best SEA-centric model (Sahabat-AI-v1 Instruct) with zero shot using BLEU, BERTScore, and LLM-as-a-judge metrics.}
\label{tab:mt-indo-local-by-region-0shot}
\end{table*}

\setlength{\tabcolsep}{0.35em}
\begin{table*}[t]
\centering
\small
\begin{tabular}{l cccc ccc ccccc}
\toprule
\multirow{2}{*}{\textbf{Province}}
& \multicolumn{4}{c}{\textbf{BLEU Scores}}
& \multicolumn{3}{c}{\textbf{BERTScore}}
& \multicolumn{5}{c}{\textbf{LLM as a judge (1--5)}} \\
\cmidrule(lr){2-5} \cmidrule(lr){6-8} \cmidrule(lr){9-13}
& \textbf{B1} & \textbf{B2} & \textbf{B3} & \textbf{B4}
& \textbf{P} & \textbf{R} & \textbf{F1}
& \textbf{adequacy} & \textbf{fluency} & \textbf{register} & \textbf{terminology} & \textbf{overall} \\
\midrule

\multicolumn{13}{c}{\textbf{Gemma-2-9B Instruct (SFT)}} \\
Aceh 
& 0.34 & 0.22 & 0.14 & 0.09 & 0.75 & 0.75 & 0.75 & 2.32 & 1.64 & 1.74 & 2.38 & 2.09 \\
Bali 
& 0.24 & 0.15 & 0.09 & 0.05 & 0.73 & 0.72 & 0.72 & 2.64 & 1.89 & 1.85 & 2.62 & 2.26 \\
West Java 
& 0.51 & 0.39 & 0.29 & 0.20 & 0.84 & 0.83 & 0.83 & 3.23 & 2.48 & 2.37 & 3.12 & 2.71 \\
Central Java 
& 0.50 & 0.37 & 0.27 & 0.19 & 0.83 & 0.82 & 0.82 & 3.25 & 2.53 & 2.44 & 3.22 & 2.80 \\
East Java 
& 0.46 & 0.32 & 0.22 & 0.15 & 0.82 & 0.82 & 0.82 & 3.25 & 2.54 & 2.44 & 3.31 & 2.75 \\
South Kalimantan
& 0.53 & 0.39 & 0.28 & 0.20 & 0.83 & 0.81 & 0.82 & 3.75 & 2.64 & 2.26 & 3.58 & 2.70 \\
NTT 
& 0.04 & 0.02 & 0.01 & 0.01 & 0.57 & 0.57 & 0.57 & 1.40 & 1.07 & 1.07 & 1.40 & 1.27 \\
Papua 
& 0.11 & 0.06 & 0.03 & 0.02 & 0.62 & 0.61 & 0.62 & 1.31 & 1.11 & 1.03 & 1.24 & 1.12 \\
South Sulawesi 
& 0.05 & 0.03 & 0.01 & 0.01 & 0.58 & 0.57 & 0.57 & 1.57 & 1.22 & 1.22 & 1.52 & 1.37 \\
West Sumatera 
& \textbf{0.55} & \textbf{0.41} & \textbf{0.30} & \textbf{0.22} & \textbf{0.85} & \textbf{0.84} & \textbf{0.84} & \textbf{4.23} & \textbf{3.45} & \textbf{3.47} & \textbf{4.26} & \textbf{3.54} \\

\midrule
\multicolumn{13}{c}{\textbf{Sahabat-AI-v1 Instruct (SFT)}} \\
Aceh 
& 0.37 & 0.24 & 0.16 & 0.10 & 0.78 & 0.76 & 0.77 & 2.37 & 1.76 & 1.86 & 2.43 & 2.14 \\
Bali 
& 0.27 & 0.16 & 0.10 & 0.06 & 0.74 & 0.73 & 0.73 & 2.60 & 1.98 & 1.99 & 2.64 & 2.35 \\
West Java 
& 0.52 & 0.38 & 0.28 & 0.19 & 0.85 & 0.84 & 0.85 & 4.42 & 3.76 & 3.69 & 4.42 & 3.80 \\
Central Java 
& \textbf{0.57} & \textbf{0.45} & \textbf{0.35} & \textbf{0.27} & \textbf{0.86} & 0.84 & 0.85 & \textbf{4.53} & 3.90 & 3.80 & 4.51 & 3.92 \\
East Java 
& 0.55 & 0.42 & 0.31 & 0.23 & 0.85 & 0.84 & 0.85 & 4.49 & \textbf{4.03} & \textbf{3.92} & \textbf{4.56} & \textbf{4.03} \\
South Kalimantan
& 0.54 & 0.40 & 0.30 & 0.21 & 0.84 & 0.82 & 0.83 & 4.13 & 3.28 & 2.99 & 4.08 & 3.23 \\
NTT 
& 0.04 & 0.02 & 0.01 & 0.00 & 0.59 & 0.60 & 0.59 & 1.33 & 1.13 & 1.13 & 1.40 & 1.27 \\
Papua 
& 0.09 & 0.05 & 0.02 & 0.01 & 0.59 & 0.60 & 0.59 & 1.46 & 1.22 & 1.21 & 1.40 & 1.30 \\
South Sulawesi 
& 0.05 & 0.03 & 0.01 & 0.01 & 0.57 & 0.57 & 0.57 & 1.51 & 1.19 & 1.21 & 1.51 & 1.30 \\
West Sumatera 
& \textbf{0.57} & 0.44 & 0.34 & 0.25 & \textbf{0.86} & \textbf{0.85} & \textbf{0.86} & 3.92 & 3.38 & 3.42 & 4.02 & 3.49 \\

\bottomrule
\end{tabular}
\caption{Translation results for \textbf{Indonesian $\rightarrow$ local} under \textbf{SFT models}, broken down by provinces. We compare the best multilingual model (Gemma-2-9b-it) and SEA-centric model (Sahabat-AI-v1 Instruct) using BLEU, BERTScore, and LLM-as-a-judge metrics.}
\label{tab:mt-indo-local-by-region-sft}
\end{table*}

\setlength{\tabcolsep}{0.35em}
\begin{table*}[t]
\centering
\small
\begin{tabular}{l cccc ccc ccccc}
\toprule
\multirow{2}{*}{\textbf{Province}}
& \multicolumn{4}{c}{\textbf{BLEU Scores}}
& \multicolumn{3}{c}{\textbf{BERTScore}}
& \multicolumn{5}{c}{\textbf{LLM as a judge (1--5)}} \\
\cmidrule(lr){2-5} \cmidrule(lr){6-8} \cmidrule(lr){9-13}
& \textbf{B1} & \textbf{B2} & \textbf{B3} & \textbf{B4}
& \textbf{P} & \textbf{R} & \textbf{F1}
& \textbf{adequacy} & \textbf{fluency} & \textbf{register} & \textbf{terminology} & \textbf{overall} \\
\midrule

\multicolumn{13}{c}{\textbf{Gemini-3-Flash (reasoning none)}} \\
Aceh 
& 0.76 & 0.67 & 0.59 & 0.51 & 0.93 & 0.92 & 0.92 & 4.50 & 4.96 & 4.74 & 4.66 & 4.58 \\
Bali 
& 0.69 & 0.58 & 0.50 & 0.42 & 0.91 & 0.91 & 0.91 & 4.41 & 4.98 & 4.80 & 4.58 & 4.48 \\
West Java 
& \textbf{0.82} & \textbf{0.75} & \textbf{0.69} & \textbf{0.63} & \textbf{0.94} & \textbf{0.93} & \textbf{0.94} & 4.83 & 4.97 & 4.88 & 4.91 & 4.87 \\
Central Java 
& 0.78 & 0.70 & 0.63 & 0.57 & 0.93 & \textbf{0.93} & 0.93 & 4.92 & 4.97 & 4.91 & \textbf{4.94} & \textbf{4.93} \\
East Java 
& 0.73 & 0.63 & 0.54 & 0.46 & 0.91 & 0.91 & 0.91 & \textbf{4.93} & 4.95 & 4.89 & \textbf{4.94} & 4.90 \\
South Kalimantan
& 0.80 & 0.72 & 0.65 & 0.59 & 0.93 & 0.92 & 0.93 & 4.80 & 4.97 & \textbf{4.92} & 4.85 & 4.84 \\
NTT 
& 0.39 & 0.25 & 0.16 & 0.10 & 0.83 & 0.82 & 0.82 & 3.53 & 4.53 & 4.27 & 3.53 & 3.53 \\
Papua 
& 0.35 & 0.21 & 0.13 & 0.09 & 0.80 & 0.79 & 0.79 & 3.55 & 4.93 & 4.68 & 3.82 & 3.86 \\
South Sulawesi
& 0.61 & 0.48 & 0.38 & 0.30 & 0.87 & 0.87 & 0.87 & 3.88 & 4.96 & 4.78 & 3.88 & 3.93 \\
West Sumatera
& 0.78 & 0.71 & 0.64 & 0.58 & 0.93 & 0.92 & 0.92 & 4.89 & \textbf{4.99} & 4.91 & 4.86 & 4.91 \\

\midrule
\multicolumn{13}{c}{\textbf{Gemma-2-9B Instruct}} \\
Aceh 
& 0.50 & 0.37 & 0.27 & 0.20 & 0.85 & 0.83 & 0.84 & 2.87 & 4.13 & 3.60 & 3.18 & 3.23 \\
Bali 
& 0.46 & 0.33 & 0.24 & 0.17 & 0.83 & 0.83 & 0.83 & 2.65 & 3.69 & 3.14 & 2.81 & 2.88 \\
West Java 
& 0.68 & 0.57 & 0.48 & 0.40 & 0.89 & 0.88 & 0.89 & 3.40 & 4.19 & 3.85 & 3.67 & 3.56 \\
Central Java 
& \textbf{0.71} & \textbf{0.61} & \textbf{0.52} & \textbf{0.45} & \textbf{0.91} & \textbf{0.90} & \textbf{0.90} & \textbf{4.18} & \textbf{4.64} & \textbf{4.41} & \textbf{4.43} & \textbf{4.26} \\
East Java 
& 0.66 & 0.54 & 0.44 & 0.36 & 0.89 & 0.88 & 0.88 & 3.89 & 4.42 & 4.08 & 4.10 & 3.92 \\
South Kalimantan
& 0.67 & 0.56 & 0.47 & 0.40 & 0.89 & 0.88 & 0.88 & 3.71 & 4.37 & 4.11 & 3.89 & 3.81 \\
NTT 
& 0.20 & 0.10 & 0.04 & 0.02 & 0.76 & 0.73 & 0.74 & 2.00 & 3.60 & 3.07 & 2.20 & 2.53 \\
Papua 
& 0.25 & 0.13 & 0.07 & 0.04 & 0.76 & 0.75 & 0.76 & 2.41 & 4.12 & 3.57 & 2.61 & 2.89 \\
South Sulawesi
& 0.28 & 0.16 & 0.09 & 0.05 & 0.76 & 0.76 & 0.76 & 1.46 & 3.37 & 2.52 & 1.72 & 2.00 \\
West Sumatera
& 0.70 & 0.60 & \textbf{0.52} & \textbf{0.45} & 0.90 & 0.88 & 0.89 & 3.84 & 4.43 & 4.22 & 3.98 & 3.91 \\

\midrule
\multicolumn{13}{c}{\textbf{Sahabat-AI-v1 Instruct}} \\
Aceh 
& 0.53 & 0.39 & 0.30 & 0.22 & 0.85 & 0.84 & 0.85 & 3.23 & 4.51 & 4.03 & 3.48 & 3.51 \\
Bali 
& 0.48 & 0.34 & 0.25 & 0.18 & 0.84 & 0.83 & 0.83 & 2.86 & 4.09 & 3.49 & 3.03 & 3.17 \\
West Java 
& 0.74 & 0.65 & 0.57 & 0.49 & 0.91 & 0.90 & 0.91 & 4.17 & 4.64 & 4.44 & 4.31 & 4.25 \\
Central Java 
& \textbf{0.75} & \textbf{0.66} & \textbf{0.58} & \textbf{0.51} & \textbf{0.92} & \textbf{0.91} & \textbf{0.92} & \textbf{4.65} & \textbf{4.90} & \textbf{4.76} & \textbf{4.74} & \textbf{4.67} \\
East Java 
& 0.70 & 0.59 & 0.50 & 0.42 & 0.90 & 0.89 & 0.90 & 4.50 & 4.82 & 4.64 & 4.63 & 4.54 \\
South Kalimantan
& 0.71 & 0.61 & 0.52 & 0.44 & 0.90 & 0.89 & 0.90 & 4.12 & 4.73 & 4.52 & 4.24 & 4.22 \\
NTT 
& 0.22 & 0.11 & 0.05 & 0.03 & 0.75 & 0.73 & 0.74 & 2.13 & 4.13 & 3.53 & 2.53 & 2.80 \\
Papua 
& 0.25 & 0.13 & 0.07 & 0.04 & 0.76 & 0.75 & 0.76 & 2.64 & 4.47 & 3.83 & 2.92 & 3.14 \\
South Sulawesi 
& 0.28 & 0.16 & 0.09 & 0.05 & 0.76 & 0.76 & 0.76 & 1.49 & 3.70 & 2.79 & 1.75 & 2.15 \\
West Sumatera 
& 0.73 & 0.63 & 0.55 & 0.48 & 0.91 & 0.90 & 0.90 & 4.25 & 4.79 & 4.57 & 4.37 & 4.34 \\

\bottomrule
\end{tabular}
\caption{Translation results for \textbf{local $\rightarrow$ Indonesian} under the \textbf{0-shot setting}, broken down by provinces. We compare the best proprietary, multilingual, and SEA-centric models using BLEU, BERTScore, and LLM-as-a-judge metrics.}
\label{tab:mt-local-indo-by-region-0shot}
\end{table*}

\setlength{\tabcolsep}{0.35em}
\begin{table*}[t]
\centering
\small
\begin{tabular}{l cccc ccc ccccc}
\toprule
\multirow{2}{*}{\textbf{Province}}
& \multicolumn{4}{c}{\textbf{BLEU Scores}}
& \multicolumn{3}{c}{\textbf{BERTScore}}
& \multicolumn{5}{c}{\textbf{LLM as a judge (1--5)}} \\
\cmidrule(lr){2-5} \cmidrule(lr){6-8} \cmidrule(lr){9-13}
& \textbf{B1} & \textbf{B2} & \textbf{B3} & \textbf{B4}
& \textbf{P} & \textbf{R} & \textbf{F1}
& \textbf{adequacy} & \textbf{fluency} & \textbf{register} & \textbf{terminology} & \textbf{overall} \\
\midrule

\multicolumn{13}{c}{\textbf{Gemma-2-9B Instruct (SFT)}} \\
Aceh 
& 0.60 & 0.48 & 0.38 & 0.31 & 0.88 & 0.87 & 0.87 & 3.29 & 4.87 & 4.25 & 3.53 & 3.60 \\
Bali 
& 0.57 & 0.44 & 0.35 & 0.27 & 0.86 & 0.85 & 0.85 & 2.94 & 4.65 & 3.81 & 3.10 & 3.30 \\
West Java 
& 0.76 & 0.67 & 0.60 & 0.54 & 0.92 & 0.91 & \textbf{0.92} & 3.81 & 4.74 & 4.34 & 4.04 & 4.00 \\
Central Java 
& \textbf{0.77} & \textbf{0.69} & \textbf{0.62} & \textbf{0.55} & \textbf{0.93} & \textbf{0.92} & \textbf{0.92} & \textbf{4.23} & 4.81 & \textbf{4.47} & \textbf{4.48} & \textbf{4.35} \\
East Java 
& 0.72 & 0.61 & 0.52 & 0.44 & 0.91 & 0.90 & 0.91 & 3.94 & 4.75 & 4.33 & 4.14 & 4.07 \\
South Kalimantan
& 0.76 & 0.67 & 0.60 & 0.53 & 0.92 & 0.91 & \textbf{0.92} & 4.02 & \textbf{4.93} & 4.46 & 4.20 & 4.18 \\
NTT 
& 0.31 & 0.17 & 0.08 & 0.04 & 0.79 & 0.77 & 0.78 & 2.13 & 4.73 & 3.80 & 2.27 & 2.87 \\
Papua 
& 0.33 & 0.19 & 0.11 & 0.07 & 0.80 & 0.78 & 0.79 & 2.61 & 4.79 & 4.00 & 2.88 & 3.18 \\
South Sulawesi 
& 0.35 & 0.21 & 0.11 & 0.06 & 0.80 & 0.78 & 0.79 & 1.70 & 4.55 & 3.51 & 1.97 & 2.61 \\
West Sumatera 
& \textbf{0.77} & 0.68 & 0.61 & \textbf{0.55} & 0.92 & 0.91 & \textbf{0.92} & 4.12 & 4.84 & 4.41 & 4.22 & 4.21 \\

\midrule
\multicolumn{13}{c}{\textbf{Sahabat-AI-v1 Instruct (SFT)}} \\
Aceh 
& 0.62 & 0.50 & 0.42 & 0.34 & 0.88 & 0.87 & 0.88 & 3.50 & 4.90 & 4.34 & 3.68 & 3.75 \\
Bali 
& 0.59 & 0.47 & 0.38 & 0.30 & 0.87 & 0.86 & 0.86 & 3.23 & 4.82 & 4.04 & 3.38 & 3.52 \\
West Java 
& \textbf{0.82} & \textbf{0.75} & \textbf{0.69} & \textbf{0.63} & \textbf{0.94} & \textbf{0.93} & \textbf{0.94} & 4.31 & 4.95 & 4.64 & 4.52 & 4.47 \\
Central Java 
& 0.81 & 0.73 & 0.67 & 0.61 & \textbf{0.94} & \textbf{0.93} & \textbf{0.94} & \textbf{4.69} & \textbf{4.98} & \textbf{4.75} & \textbf{4.80} & \textbf{4.77} \\
East Java 
& 0.76 & 0.66 & 0.58 & 0.50 & 0.92 & 0.92 & 0.92 & 4.61 & 4.91 & 4.71 & 4.70 & 4.67 \\
South Kalimantan
& 0.80 & 0.71 & 0.65 & 0.59 & 0.93 & 0.92 & 0.93 & 4.28 & 4.96 & 4.65 & 4.47 & 4.41 \\
NTT 
& 0.29 & 0.17 & 0.10 & 0.06 & 0.78 & 0.76 & 0.77 & 2.00 & 4.80 & 3.60 & 2.40 & 2.87 \\
Papua 
& 0.34 & 0.21 & 0.12 & 0.07 & 0.80 & 0.79 & 0.79 & 2.75 & 4.89 & 4.15 & 2.98 & 3.30 \\
South Sulawesi 
& 0.39 & 0.25 & 0.15 & 0.09 & 0.80 & 0.79 & 0.80 & 2.03 & 4.76 & 3.78 & 2.30 & 2.82 \\
West Sumatera 
& 0.79 & 0.72 & 0.65 & 0.59 & 0.93 & \textbf{0.93} & 0.93 & 4.46 & 4.97 & 4.62 & 4.57 & 4.55 \\

\bottomrule
\end{tabular}
\caption{Translation results for \textbf{local $\rightarrow$ Indonesian} under \textbf{SFT models}, broken down by provinces. We compare the best multilingual model (Gemma-2-9b-it) and SEA-centric model (Sahabat-AI-v1 Instruct) using BLEU, BERTScore, and LLM-as-a-judge metrics.}
\label{tab:mt-local-indo-by-region-sft}
\end{table*}

\section{LLM-as-a-judge -- Human Evaluation Correlation}
\label{sec:appendix-llm-correlation}

The correlation between LLM-as-a-judge and human evaluation is presented in Table~\ref{tab:mt-llm-human-correlation-adequacy} for adequacy and Table~\ref{tab:mt-llm-human-correlation-fluency} for fluency. Across both translation directions, proprietary models consistently achieve mean absolute distance (MAD) scores below 1 while maintaining Acc@1 scores above 0.84, indicating strong agreement with human judgments. Open-source models also demonstrate reasonably good correlations, although weaker than proprietary systems. Gemma-2-9B Instruct achieves average MAD scores of 0.98 and 0.80 for adequacy and fluency, respectively, while Sailor2-8B Chat obtains 0.90 and 0.87. Their Acc@1 scores remain relatively high at 0.78 for Gemma-2-9B Instruct and 0.775 for Sailor2-8B Chat, suggesting that LLM-as-a-judge remains a reliable proxy for human evaluation even in open-source settings.

\begin{table*}[t]
\centering
\small
\begin{tabular}{lcc|cc|cc}
\toprule
\multicolumn{7}{c}{\textbf{Indonesian $\rightarrow$ Local}} \\
\midrule
 & \multicolumn{2}{c}{Gemini-3-flash-preview} & \multicolumn{2}{c}{Gemma-2-9B Instruct} & \multicolumn{2}{c}{Sailor2-8B Chat} \\
\cmidrule(lr){2-3} \cmidrule(lr){4-5} \cmidrule(lr){6-7}
 & MAD & Acc@1 & MAD & Acc@1 & MAD & Acc@1 \\
\midrule
Indonesian → Javanese  & 0.52 & 0.84 & 0.76 & 0.84 & 0.72 & 0.84 \\
Indonesian → Minangkabau  & 0.28 & 0.96 & 1.08 & 0.71 & 1.00 & 0.72 \\
\midrule
\multicolumn{7}{c}{\textbf{Local $\rightarrow$ Indonesian}} \\
\midrule
 & \multicolumn{2}{c}{GPT-5.1} & \multicolumn{2}{c}{Gemma-2-9B Instruct} & \multicolumn{2}{c}{Sahabat-AI-v1 Instruct} \\
\cmidrule(lr){2-3} \cmidrule(lr){4-5} \cmidrule(lr){6-7}
 & MAD & Acc@1 & MAD & Acc@1 & MAD & Acc@1 \\
\midrule
Javanese → Indonesian   & 0.12 & 0.96 & 1.24 & 0.68 & 0.64 & 0.76 \\
Minangkabau → Indonesian  & 0.04 & 1.00 & 0.83 & 0.79 & 1.24 & 0.76 \\
\bottomrule
\end{tabular}
\caption{MAD and Acc@1 scores between LLM-as-a-judge and human evaluation in ``adequacy'' for Indonesian $\leftrightarrow$ local translation across models.}
\label{tab:mt-llm-human-correlation-adequacy}
\end{table*}

\begin{table*}[t]
\centering
\small
\begin{tabular}{lcc|cc|cc}
\toprule
\multicolumn{7}{c}{\textbf{Indonesian $\rightarrow$ Local}} \\
\midrule
 & \multicolumn{2}{c}{Gemini-3-flash-preview} & \multicolumn{2}{c}{Gemma-2-9B Instruct} & \multicolumn{2}{c}{Sailor2-8B Chat} \\
\cmidrule(lr){2-3} \cmidrule(lr){4-5} \cmidrule(lr){6-7}
 & MAD & Acc@1 & MAD & Acc@1 & MAD & Acc@1 \\
\midrule
Indonesian → Javanese  & 0.80 & 0.92 & 0.80 & 0.84 & 1.12 & 0.76 \\
Indonesian → Minangkabau  & 0.80 & 0.92 & 0.96 & 0.83 & 1.40 & 0.56 \\
\midrule
\multicolumn{7}{c}{\textbf{Local $\rightarrow$ Indonesian}} \\
\midrule
 & \multicolumn{2}{c}{GPT-5.1} & \multicolumn{2}{c}{Gemma-2-9B Instruct} & \multicolumn{2}{c}{Sahabat-AI-v1 Instruct} \\
\cmidrule(lr){2-3} \cmidrule(lr){4-5} \cmidrule(lr){6-7}
 & MAD & Acc@1 & MAD & Acc@1 & MAD & Acc@1 \\
\midrule
Javanese → Indonesian   & 0.40 & 0.96 & 0.92 & 0.76 & 0.60 & 0.88 \\
Minangkabau → Indonesian  & 0.16 & 0.92 & 0.50 & 0.88 & 0.36 & 0.92 \\
\bottomrule
\end{tabular}
\caption{MAD and Acc@1 scores between LLM-as-a-judge and human evaluation in ``fluency'' for Indonesian $\leftrightarrow$ local translation across models.}
\label{tab:mt-llm-human-correlation-fluency}
\end{table*}

\section{Language Steering Evaluation Details}
\label{sec:appendix-dialogue-steering-evaluation-details}

As shown in Table~\ref{tab:appendix-dialogue-steering}, models generally perform well when generating Indonesian utterances. Even the lowest-performing model (Gemma-2-9B Instruct) produces Indonesian outputs correctly in 91.3\% of cases, as measured by the Glot-LID metric. GPT-5.1 achieves the best performance, with an LLM-as-a-judge score of 4.65 on a 5-point Likert scale.

Consistent with the main results, model performance drops substantially when generating Indonesian local languages. For example, Sahabat-AI-v1 Instruct, which achieves an LLM-as-a-judge score of 4.15 and a 95.0\% correct language generation rate in Indonesian, drops to 3.32 and 13.7\%, respectively, in local language generation. We also observe that SFT improves the ability of open-source models to generate the correct local languages, although this comes with a slight degradation in Indonesian generation performance.

We further provide province-level results in Table~\ref{tab:dialogue-steering-region}. Overall, even strong proprietary models such as Gemini-3-Flash still struggle with languages spoken in Eastern Indonesia, including those from NTT, South Sulawesi, and Papua. In a good sign, models achieve relatively strong LLM-as-a-judge scores despite generating responses primarily in Indonesian rather than the target local language. It is also worth noting that the GLOT-LID score for Papua (Wamesa, `\textit{wad}') is reported as `\textit{NA}' because Wamesa is not covered in the GLOT-LID label set.

\setlength{\tabcolsep}{0.05em}
\begin{table*}[t]
\centering
\footnotesize
\begin{tabular}{l cccc cccc}
\toprule
\multirow{2}{*}{\textbf{Model}}
& \multicolumn{4}{c}{\textbf{Local $\rightarrow$ Indonesian}}
& \multicolumn{4}{c}{\textbf{Indonesian $\rightarrow$ Local}} \\
\cmidrule(lr){2-5} \cmidrule(lr){6-9}
& \textbf{LLM} & \textbf{Glot-LID} & \textbf{BLEU} & \textbf{chrF}
& \textbf{LLM} & \textbf{Glot-LID} & \textbf{BLEU} & \textbf{chrF} \\
\midrule

\multicolumn{9}{c}{\textbf{Proprietary Models}} \\
GPT-5.1 
& \textbf{4.65} & \textbf{98.3} & 1.80 & \textbf{26.02}
& 4.42 & 73.9 & 0.64 & 20.57 \\
Gemini-3-flash-preview 
& 4.62 & 97.8 & 2.25 & 25.28
& \textbf{4.55} & \textbf{92.7} & \textbf{0.94} & \textbf{20.98} \\
Cohere-Command-A 
& 4.37 & 97.9 & \textbf{2.96} & 25.23
& 3.70 & 31.6 & 0.62 & 18.35 \\
\midrule

\multicolumn{9}{c}{\textbf{Multilingual Models}} \\
Qwen3-8B 
& \textbf{3.94} \textsubscript{($\pm$0.0)} & 98.7 \textsubscript{($\pm$0.1)} & \textbf{1.75} \textsubscript{($\pm$0.0)} & 21.38 \textsubscript{($\pm$0.0)}
& \textbf{3.11} \textsubscript{($\pm$0.0)} & 1.5 \textsubscript{($\pm$0.2)} & 0.42 \textsubscript{($\pm$0.1)} & 15.79 \textsubscript{($\pm$0.0)} \\
Llama-3.1-8B Instruct 
& 3.80 \textsubscript{($\pm$0.0)} & \textbf{99.4} \textsubscript{($\pm$0.1)} & 1.34 \textsubscript{($\pm$0.1)} & \textbf{22.67} \textsubscript{($\pm$0.1)}
& 2.91 \textsubscript{($\pm$0.0)} & \textbf{9.2} \textsubscript{($\pm$0.4)} & 0.32 \textsubscript{($\pm$0.2)} & \textbf{16.68} \textsubscript{($\pm$0.1)} \\
Gemma-2-9b Instruct 
& 3.78 \textsubscript{($\pm$0.0)} & 91.3 \textsubscript{($\pm$0.5)} & 1.62 \textsubscript{($\pm$0.1)} & 14.96 \textsubscript{($\pm$0.0)}
& 3.00 \textsubscript{($\pm$0.0)} & 7.8 \textsubscript{($\pm$0.6)} & \textbf{0.44} \textsubscript{($\pm$0.0)} & 10.92 \textsubscript{($\pm$0.1)} \\
\midrule

\multicolumn{9}{c}{\textbf{SEA-Centric Models}} \\
Sailor2-8B Chat 
& 3.93 \textsubscript{($\pm$0.0)} & 95.5 \textsubscript{($\pm$0.1)} & 0.67 \textsubscript{($\pm$0.0)} & 21.42 \textsubscript{($\pm$0.1)}
& 3.19 \textsubscript{($\pm$0.0)} & 23.7 \textsubscript{($\pm$0.4)} & 0.23 \textsubscript{($\pm$0.0)} & \textbf{17.43} \textsubscript{($\pm$0.1)} \\
SEA-LION-v3.5-8B 
& 3.82 \textsubscript{($\pm$0.0)} & \textbf{99.1} \textsubscript{($\pm$0.2)} & 1.22 \textsubscript{($\pm$0.0)} & \textbf{22.56} \textsubscript{($\pm$0.1)}
& 3.01 \textsubscript{($\pm$0.0)} & \textbf{27.5} \textsubscript{($\pm$0.7)} & 0.26 \textsubscript{($\pm$0.0)} & 16.64 \textsubscript{($\pm$0.1)} \\
Sahabat-AI-v1 Instruct 
& \textbf{4.15} \textsubscript{($\pm$0.0)} & 95.0 \textsubscript{($\pm$0.5)} & \textbf{3.13} \textsubscript{($\pm$0.1)} & 20.58 \textsubscript{($\pm$0.1)}
& \textbf{3.32} \textsubscript{($\pm$0.0)} & 13.7 \textsubscript{($\pm$1.0)} & \textbf{0.64} \textsubscript{($\pm$0.2)} & 13.65 \textsubscript{($\pm$0.0)} \\
\midrule

\multicolumn{9}{c}{\textbf{SFT Models}} \\
Qwen3-8B 
& \red{$\Downarrow$} 3.84 \textsubscript{($\pm$0.0)} & \red{$\Downarrow$} \textbf{96.0} \textsubscript{($\pm$0.0)} & \green{$\Uparrow$} \textbf{3.79} \textsubscript{($\pm$0.0)} & \green{$\Uparrow$} \textbf{25.12} \textsubscript{($\pm$0.2)}
& \red{$\Downarrow$} 3.06 \textsubscript{($\pm$0.1)} & \green{$\Uparrow$} 4.4 \textsubscript{($\pm$5.1)} & \green{$\Uparrow$} 0.50 \textsubscript{($\pm$0.1)} & \green{$\Uparrow$} \textbf{16.18} \textsubscript{($\pm$0.6)} \\
Sahabat-AI-v1 Instruct 
& \red{$\Downarrow$} \textbf{4.12} \textsubscript{($\pm$0.0)} & \red{$\Downarrow$} 94.6 \textsubscript{($\pm$0.0)} & \red{$\Downarrow$} 3.08 \textsubscript{($\pm$0.1)} & \red{$\Downarrow$} 20.47 \textsubscript{($\pm$0.1)}
& \green{$\Uparrow$} \textbf{3.33} \textsubscript{($\pm$0.1)} & \green{$\Uparrow$} \textbf{18.2} \textsubscript{($\pm$8.9)} & \green{$\Uparrow$} \textbf{0.77} \textsubscript{($\pm$0.0)} & \green{$\Uparrow$} 14.74 \textsubscript{($\pm$1.8)} \\
\bottomrule
\end{tabular}
\caption{Language steering results in \textbf{Indonesian} and \textbf{Local languages}. We report LLM-as-a-judge (denoted as LLM), Glot-LID, BLEU, and chrF scores.}
\label{tab:appendix-dialogue-steering}
\end{table*}

\setlength{\tabcolsep}{0.25em}
\begin{table*}[t]
\centering
\small
\begin{tabular}{l cccc cccc cccc}
\toprule
\multirow{2}{*}{\textbf{Province}}
& \multicolumn{4}{c}{\textbf{Gemini-3-Flash}}
& \multicolumn{4}{c}{\textbf{Qwen3-8b}}
& \multicolumn{4}{c}{\textbf{SEA-LION-v3.5-8B}} \\
\cmidrule(lr){2-5} \cmidrule(lr){6-9} \cmidrule(lr){10-13}
& \textbf{LLM} & \textbf{Glot-LID} & \textbf{BLEU} & \textbf{chrF}
& \textbf{LLM} & \textbf{Glot-LID} & \textbf{BLEU} & \textbf{chrF}
& \textbf{LLM} & \textbf{Glot-LID} & \textbf{BLEU} & \textbf{chrF} \\
\midrule

Aceh 
& 4.62 & 99.50 & 0.42 & 20.03 
& 3.04 & 0.00 & 0.45 & 14.66 
& 3.01 & 1.51 & 0.31 & 11.63 \\

Bali 
& 4.48 & 94.50 & 0.36 & 21.35 
& 3.17 & 0.00 & 0.38 & 16.37 
& 3.21 & 2.00 & 0.52 & 14.10 \\

West Java
& 4.43 & 92.79 & 0.88 & 21.54 
& 2.98 & \textbf{6.60} & 0.57 & 16.38 
& \textbf{3.54} & \textbf{40.10} & 1.04 & 15.26 \\

Central Java 
& 4.67 & 99.00 & 1.29 & 21.01 
& 3.10 & 1.30 & 0.57 & 15.85 
& 3.48 & 21.90 & 0.67 & 13.87 \\

East Java
& 4.61 & 98.70 & 1.00 & 23.07 
& 3.16 & 2.60 & \textbf{0.89} & 16.52 
& 3.53 & 23.59 & \textbf{1.55} & 15.54 \\

South Kalimantan 
& 4.60 & 99.00 & 0.53 & 24.48 
& 3.12 & 1.10 & 0.47 & \textbf{18.15} 
& 3.27 & 6.35 & 0.90 & \textbf{15.82} \\

NTT 
& 4.45 & 0.00 & 0.37 & 14.53 
& \textbf{3.26} & 0.00 & 0.68 & 12.72 
& 3.26 & 0.00 & 0.89 & 10.95 \\

Papua 
& 4.42 & \textit{NA} & 0.29 & 15.65 
& 3.25 & \textit{NA} & 0.29 & 13.28 
& 3.26 & \textit{NA} & 0.33 & 10.05 \\

South Sulawesi
& 4.15 & 3.00 & 1.10 & 17.55 
& 3.00 & 0.00 & 0.09 & 13.05 
& 3.18 & 0.00 & 0.28 & 10.50 \\

West Sumatera
& \textbf{4.68} & \textbf{100.00} & \textbf{1.88} & \textbf{24.73} 
& 3.11 & 0.50 & 0.60 & 17.74 
& 3.38 & 10.78 & 1.05 & 16.4 \\

\bottomrule
\end{tabular}
\caption{Language steering results across provinces comparing the best proprietary, multilingual, and SEA-centric models in this task. We report LLM-as-a-judge (denoted as LLM), Glot-LID, BLEU, and chrF scores.}
\label{tab:dialogue-steering-region}
\end{table*}

\end{document}